\documentclass[twoside,11pt]{article}

\usepackage{jmlr2e}

\usepackage{amsmath}
\usepackage{url}% simple URL typesetting
\usepackage{amsfonts}% blackboard math symbols
\usepackage{nicefrac}% compact symbols for 1/2, etc.
\usepackage{microtype}% microtypography
\usepackage{bm}
\usepackage{multirow}
\usepackage{array}
\usepackage{rotating}
\usepackage{float}
\usepackage{wrapfig}
\usepackage{color}
\usepackage{array}
\usepackage{booktabs}
\usepackage{bm}
\usepackage{multirow}
\usepackage{amsmath}
\usepackage{amssymb}
\usepackage{graphicx}

\usepackage{subfigure}
\usepackage{times}

\usepackage{algorithm}\usepackage{algorithmic}% TeX will automatically convert eps --> pdf in pdflatex		
\usepackage{bm}

\ShortHeadings{$S^{2}$-LBI: Stochastic Split Linearized Bregman Iterations for Parsimonious Deep Learning}{The Technical Report on training deep networks by Split LBI}
\firstpageno{1}

\begin{document}

\title{$S^{2}$-LBI: Stochastic Split Linearized Bregman Iterations for Parsimonious Deep Learning }

\author{\name Yanwei Fu \email yanweifu@fudan.edu.cn \\
       \addr School of Data Science\\
       Fudan University\\
       Shanghai, 200433, China
    \AND
        \name Donghao Li \email 15307100013@fudan.edu.cn \\
       \addr School of Data Science\\
       Fudan University\\
       Shanghai, 200433, China
       \AND
\name Xinwei Sun \email sxwxiaoxiaohehe@pku.edu.cn \\
       \addr School of Mathematical Science\\
       Peking University\\
       Beijing, 100871, China
       \AND
       \name Shun Zhang \email 15300180012@fudan.edu.cn \\
 \addr School of Data Science\\
       Fudan University\\
       Shanghai, 200433, China
       \AND
   \name Yizhou Wang \email yizhou.wang@pku.edu.cn \\
       \addr Nat'l Engineering Laboratory for Video Technology;\\
        Key Laboratory of Machine Perception (MoE);\\
        Cooperative Medianet Innovation Center, Shanghai;\\
        School of EECS\\
       Peking University\\
       Beijing, 100871, China
           \AND
       \name Yuan Yao$^\ddagger$ \email yuany@ust.hk \\
       \addr Department of Mathematics, Chemical \& Biological Engineering;\\
        by courtesy, Computer Science \& Engineering\\
       Hong Kong University of Science and Technology\\
       Hong Kong SAR, China
       }
      \footnote{Yuan Yao$^\ddagger$ is corresponding authors.}

\editor{The Technical Report on training deep networks by Split LBI.}
\maketitle

\begin{abstract}
	This paper proposes a novel Stochastic Split Linearized Bregman Iteration
	($S^{2}$-LBI) algorithm to efficiently train the deep network. The
	$S^{2}$-LBI introduces an iterative regularization path with structural
	sparsity. Our $S^{2}$-LBI combines the computational efficiency of
	the LBI, and model selection consistency in learning the structural
	sparsity. The computed solution path intrinsically enables us to enlarge
	or simplify a network, which theoretically, is benefited from the
	dynamics property of our $S^{2}$-LBI algorithm. The experimental
	results validate our $S^{2}$-LBI on MNIST and CIFAR-10 dataset. For
	example, in MNIST, we can either boost a network with only 1.5K parameters
	(1 convolutional layer of 5 filters, and 1 FC layer), achieves 98.40\%
	recognition accuracy; or we simplify $82.5\%$ of parameters in LeNet-5
	network, and still achieves the 98.47\% recognition accuracy.   In addition, we also have the learning results on  ImageNet, which will be added in the next version of our report.
\end{abstract}

\section{Introduction}

The expressive power of Deep Convolutional Neural Networks (DNNs)
comes from the millions of parameters, which are optimized by various
algorithms such as Stochastic Gradient Descent (SGD) \cite{bottou-2010},
and Adam \cite{kingma2014adam}. Remarkably, the most popular deep
architectures are manually designed by human experts in some common
tasks, \emph{e.g.}, object categorization on ImageNet. In contrast, we
have to make a trade-off between the representation capability and
computational cost of networks in the real world applications, \emph{e.g.,}
robotics, self-driving cars, and augmented reality. On the other hand,
some experimental results show that classical DNNs may be too complicated
for most specific tasks, \emph{e.g.}, ``reducing\emph{ }$2\times$
connections without losing accuracy and without retraining'' in \cite{han2015learning}.

To explore a parsimonious deep learning structure, recent research
focuses on employing Network Architecture Search (NAS) \cite{nas_survey_2018}
and network pruning \cite{han2015deep}. Practically, the computational
cost of NAS algorithms themselves are prohibitive expensive, \emph{e.g.},
800 GPUs concurrently at any time training the algorithms in \cite{zoph2016neural}.
Furthermore, network pruning algorithms \cite{han2015deep,greedy_filter_pruning,molchanov2016pruning}
introduce additional computational cost in fine-tuning/updating the
networks. In addition, some works study the manually designed compact
and lightweight small DNNs, (\emph{e.g.} ShuffleNet \cite{shufflenetv2},
MobileNet \cite{howard2017mobilenets}, and SqueezeNet \cite{squeezenet}),
which, nevertheless, may still be tailored only for some specific
tasks. 

The additional computational cost of NAS and pruning algorithm, is due to the fact that the SGD algorithms and its variants, \emph{e.g.},
Adam, do not favor a parsimonious training with increasing levels of structural sparsity (e.g. (group) sparsity in fully connected/convolutional layer). More specifically, the SGD algorithms, in principle, do
not have an intrinsic feature selection mechanism in deep network
(Sec. \ref{subsec:Boosting-with-Sparsity}). One has to associate the loss function with some sparsity-enforcement penalties (e.g. Lasso), and for each of the regularization parameters typically in a decreasing sequence, SGD runs to seek a convergent model which is nearly never met in deep learning and then pursue various sparse models. In such a homotopy computation of Lasso-type regularization paths, various models along the SGD path have never been fully explored in any previous works.

In this paper, we propose a natural algorithm that can simultaneously follow a path of SGD training of deep learning while pursuing model structures along the path. The algorithm lifts the original network parameters, say $\Theta$, to a coupled parameter set, say $(\Theta, \Gamma)$. While a usual stochastic gradient descent runs over the primal parameter $\Theta$, a sparse proximal gradient descent (or mirror descent, linearized Bregman iterations) runs over the dual parameter $\Gamma$ which enforces structural sparsity on network models. These two dynamics are coupled using a proximal gap penalty in loss function. Then the algorithm returns an iterative path of models with multiple levels of structural sparsity. During the training process, the earlier a network structure parameters becomes non-zero, the more important is the model structure. Similarly, the larger is the magnitude of $\Gamma$, the more influence on prediction it has. Such a feature can be used to construct various network structures parsimoniously with as good a prediction performance as possible. This algorithm is inspired by a recent development of differential inclusion approach to boosting with structural sparsity that improves the Lasso path in model selection \cite{Splitlbi,Huang2017Boosting} and has been successfully used in machine vision \cite{zhao2018msplit} and neuroimage analysis \cite{sun2017gsplit}.
In this paper we adopt such an idea for the development of a novel network training algorithm, called Stochastic Split Linearized Bregman Iteration
($S^{2}$-LBI) algorithm.

\begin{figure}[ht]
	\centering{}\includegraphics[scale=0.4]{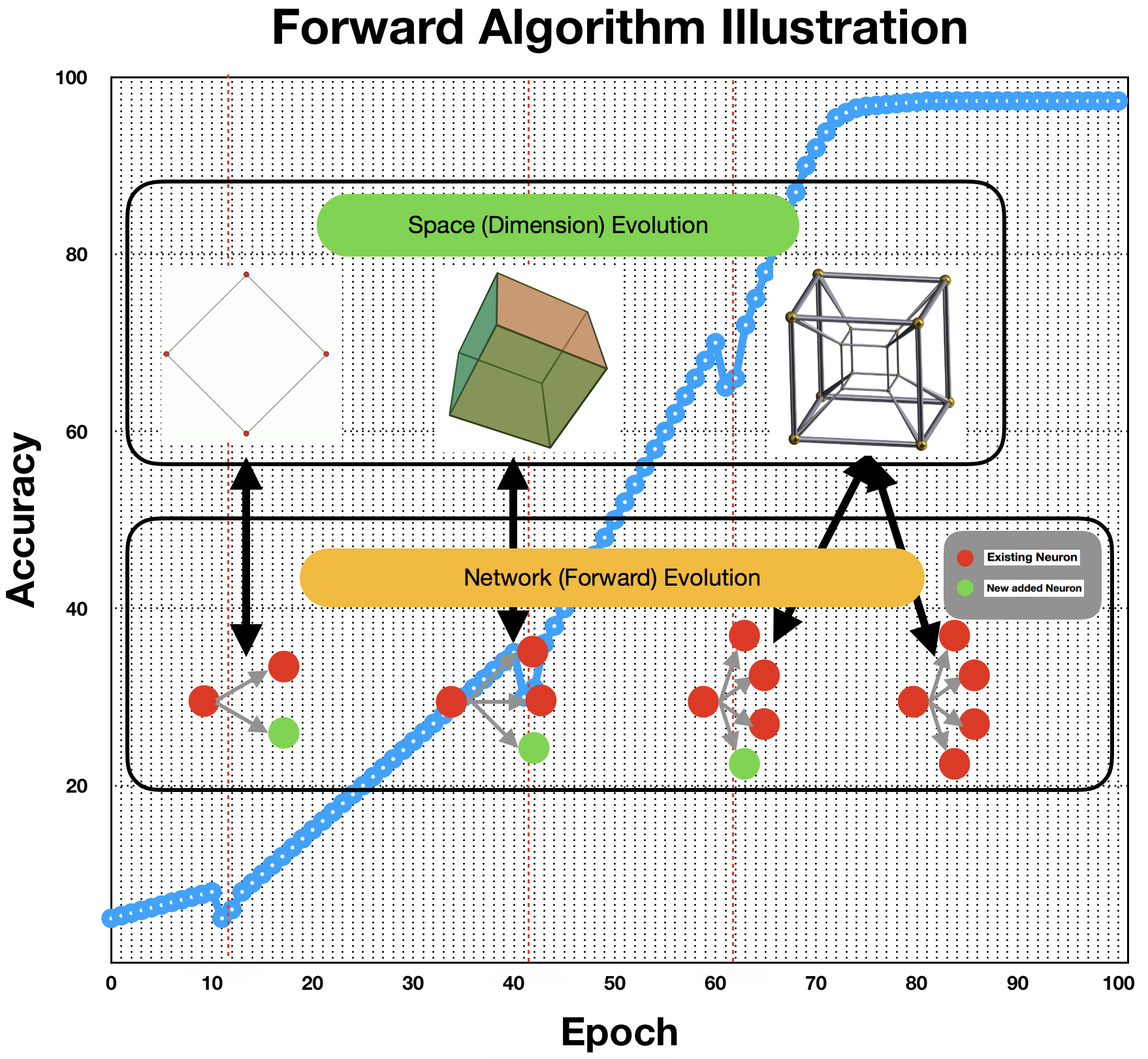}\caption{Illustration of forward algorithm. Blue represents accuracy curve, while red vertical line denotes the point at which we add new parameters. During training, the accuracy curve tends to increase with new parameters dynamically added, which lift the parameters into a higher dimensional space. When new parameters added, the accuracy decreases a little since the new random initialized parameters, and increases with fine-tuning.  \label{fig:illustration} }
\end{figure}

By virtue of solution path, we further propose
the forward/backward selection algorithms by $S^{2}$-LBI to learn
to add/remove parameters of networks.  As a result, network structures
can be ``shaped'' via forward or backward regularization
solution path computed by $S^{2}$-LBI. For the forward direction, the forward selection algorithm can dynamically expand from an initial small network into a reasonable large one, as shown in Fig.~\ref{fig:illustration} for illustration. In more details, at each stage, $S^2$-LBI can be trained to gradually fit data. Equipped with an early stopping scheme, the model can avoid overfitting in the current parameter space. Then we lift the parameter space into a higher dimensional space by adding more randomly initialized parameters (e.g. filters or neurons) and continue training again. In contrast, by configuring the
solution path in a backward manner, the backward selection algorithm can simplify or prune the initial large network to a small reasonable one. Specifically, the $S^2$-LBI can efficiently learn the importance of the parameters by measuring their selection order and magnitudes. By ranking the importance of parameters into a descending order and remove the low rank parameters, the network can thus be shrunken without losing much representation (thus prediction) power. 

\textbf{Contributions.} (1) We,
for the first time, propose a novel $S^{2}$-LBI algorithm that can
efficiently compute the regularization solution path of deep networks.
$S^{2}$-LBI algorithm can efficiently train a network with structural
sparsity. (2) In the forward manner, our forward selection algorithm
can dynamically and incrementally expand from a small initial network into  a large network one. (3) In the backward manner of solution path, our backward
selection algorithm can simplify the network without deteriorating the prediction performance.
Extensive experiments on two datasets, validate the efficacy of our
proposed algorithms.

\section{Stochastic Linearized Bregman Iteration }

In a supervised learning task, we denote $f_{\Theta}: \mathbf{\mathcal{X}} \to\mathbf{\mathcal{Y}}$ as the mapping function of neural network architecture with $L$ layers, with weight parameters $\Theta = \{W^1,...,W^L\}$,  from input space $\mathbf{\mathcal{X}}$ and output space $\mathbf{\mathcal{Y}}$. The $W^l$ is the weight matrix between $l$-th layer and ($l$+1)-th layer, i.e. $\mathcal{O}^{l+1} = \sigma(W^l\mathcal{O}^l)$ where $\mathcal{O}^l$ denotes the output of the $l$-th layer and $\sigma$ is the activation function (e.g. relu,sigmoid). Denote $\Theta_{-W^l}$ as $\{W^1,...,W^{l-1},W^{l+1},...,W^L\}$. Given training data $\mathbf{X} \overset{\Delta}{=} \{\mathbf{x}_1,...,\mathbf{x}_{N}\}$ and $\mathbf{Y} \overset{\Delta}{=} \{\mathbf{y}_1,...,\mathbf{y}_{N}\}$, we denote $\mathcal{L}(f_{\Theta}(\mathbf{X}),\mathbf{Y})$ as the loss function. For classification task with $\mathbf{\mathcal{Y}} = [K] \overset{\Delta}{=} \{1,...,K\}$, the $\mathcal{L}$ can be written as:
\begin{align}
\mathcal{L}(f_{\Theta}(\mathbf{X}),\mathbf{Y}) = \frac{1}{N}\sum_{i=1}^{N}\sum_{k \in [K]} \mathbf{1}(\mathbf{y}_{i} = k)\log{\frac{1}{p(f_{\Theta}(\mathbf{x}_i) = k)}}, \nonumber
\end{align}
We use $\mathcal{L}(\Theta)$ to denote the loss function for simplicity. Generally, the mini-batch SGD algorithm
update $\Theta$ as follows:
\begin{equation}
\Theta_{t+1}=\Theta_{t}-\alpha\cdot\nabla_{\Theta}L\left(\Theta_{t}\right)|_{(\mathbf{X},\mathbf{Y})_{{\mathrm{batch_t}}}}\label{eq:sgd}
\end{equation}
where $\Theta_t \overset{\Delta}{=} \{W_t^l\}_{l \in [L]}$ is the parameter value at the $t$-th step, $\alpha$ is
the learning rate, $(\mathbf{X},\mathbf{Y})_{{\mathrm{batch_t}}}$ represents the batch-sample at the $t$-th iteration. For simplicity, we omit $(\mathbf{X},\mathbf{Y})_{{\mathrm{batch_t}}}$ in the rest of the paper.

\subsection{Sparsity enforcement via $\ell_1$ regularization}
To achieve sparsity on a weight matrix at the $l$-th layer ($W^l$), one can apply $\ell_1$-based regularization $\Omega(W^l)$, which turns the loss function as
\begin{equation}
\mathcal{L}_{\lambda}(\Theta)=\mathcal{L}(\Theta)+\lambda\cdot\Omega\left(W^l\right)\label{eq:loss_function}
\end{equation}
where $\lambda$ is the regularization hyper-parameter. 
The choice of $\Omega(\cdot)$ is determined by the form of $W^l$:
\begin{itemize}
	\item When $W^l$ denotes filters at the $l$th layer, $\Omega(\cdot)$ is the group-lasso penalty \cite{yuan2006model}, which aims at group-wise sparsity. Specifically, $W^l$ can be re-denoted as $W_{\mathrm{conv}} \overset{\Delta}{=} \left[w_{\mathrm{conv}}^{1,\top},...,w_{\mathrm{conv}}^{G,\top}\right]^{\top}$ with $G$ filters. Then $\Omega(W^l) \overset{\Delta}{=} \Omega(W_{\mathrm{conv}}) = \sum_{g=1}^G \Vert w_{\mathrm{conv}}^{g} \Vert_2$, where $\Vert w_{\mathrm{conv}}^{g} \Vert_2 = \sqrt{\sum_k \left(w_{\mathrm{conv}}^{g,k} \right)}$. 
	\item When $W^l$ denotes the multiplication matrix before the $l+1$-th fully connected layer, $\Omega(\cdot)$ is lasso-type penalty. The $W^l$ is re-denoted as $W_{\mathrm{fc}}$, and $\Omega(W^l) \overset{\Delta}{=} \Omega(W_{\mathrm{fc}}) = \sum_{i,j} \vert W_{\mathrm{fc}}(i,j) \vert$.
\end{itemize}
This common $\ell_1$ regularization strategy can be adopted to prune the network, such as \cite{Liu_2017_ICCV,wen2016learning}. However, note that the regularization hyper-parameter $\lambda$, which determines the sparsity level of parameters, is the trade-off between under-fitting and over-fitting. Hence, one should choose an appropriate $\lambda$, which is often typically achieved via prediction error on cross-validation. In more details, one has to run SGD several times over a grid of $\lambda$'s, which brings out a large computational cost, especially for deep neural network. In the next section, we will introduce an iterative algorithm which can return a regularization solution path, which is called Stochastic Split Linearized Bregman Iteration ($S^{2}$-LBI).

\subsection{$S^{2}$-LBI algorithm}

Here we present the Stochastic Split Linearized Bregman Iteration algorithm, which we will see later, can efficiently learns the regularization solution path in optimizing neural networks. Rather than directly dealing with the sparsity of $W^l$, we exploit an augmented variable $\Gamma^l$ and want it to (i) be enforced with sparsity (ii) be kept close to $W^l$ via variable splitting term $\frac{1}{2\nu} \Vert W^l - \Gamma^l \Vert_{2}^2$ with hyper-parameter $\nu > 0$. Combined with such a variable splitting term, the loss function of $S^2$-LBI turns to 
%is Here we introduce a variable splitting term  achieve structural sparsity under linear model. The S  We present a Stochastic Split Linearized Bregman Iteration ($S^{2}$-LBI)
%algorithm, which efficiently learns the regularization solution paths
%in optimizing deep networks. Different from Eq \ref{eq:loss_function},
%the center idea of $S^{2}$-LBI is to introduce an augmented variable
%$\Gamma$ \cite{Splitlbi}, which is enforced sparsity and kept close
%to $W$ by the variable splitting term $\frac{1}{2\nu}\Vert\Gamma-W\Vert_{2}^{2}$,
%the parameter ($\nu>0$), 
\begin{align}
\bar{\mathcal{L}}\left(\Theta,\Gamma^l\right)=L\left(\Theta\right)+\frac{1}{2\nu}\left\Vert \Gamma^l-W^l\right\Vert _{2}^{2}\label{eq:slbi}
\end{align}
Now consider the following iterative algorithm for t=1,...,T:
%To efficiently implement the sparsity of $\Gamma$, the Linearized
%Bregman Iteration (LBI) is employed to modify the gradient function
%of SGD in Eq \ref{eq:sgd},
\begin{subequations} 
	\label{eq:slbi-show}
	\begin{align}
	\Theta_{-W^l,t+1} & =\Theta_{-W^l,t}-\alpha\cdot \nabla_{\Theta_{-W^l}}\bar{\mathcal{L}}\left(\Theta,\Gamma^l\right) \label{eq:slbi-show-a}\\
	W^l_{t+1} & = W^l_t -\kappa \cdot \alpha\cdot \nabla_{W^l}\bar{\mathcal{L}}\left(\Theta,\Gamma^l\right) \label{eq:slbi-show-a}\\
	Z^l_{t+1} & =Z^l_{t}-\alpha\cdot  \nabla_{\Gamma}\bar{\mathcal{L}}\left(\Theta,\Gamma^l\right)\label{eq:slbi-show-b}\\
	\Gamma^l_{t+1} & =\kappa\cdot\mathrm{Prox}_{\Omega}\left(Z^l_{t+1}\right)\label{eq:slbi-show-c}\\
	\widetilde{W}^l_{t+1} & = \mathrm{Proj}_{\mathrm{supp}(\Gamma^l_{t+1})}(W^l_{t+1})\label{eq:slbi-show-d}
	\end{align}
\end{subequations}
With $Z^l_0 = \Gamma^l_0 = 0$, $\Theta_0$ can be randomly initialized from Gaussian Distribution, and hyper-parameters $\nu,\kappa,\alpha > 0$. The $Z$ here is an auxiliary parameter used for gradient descent w.r.t. $\Gamma^l$, where by Moreau decomposition $Z = \Gamma + \partial(\Omega(\Gamma))/\kappa$. The sparsity of $\Gamma$ is achieved by proximal map in~\eqref{eq:slbi-show-c} associated with
$\Omega\left(\cdot\right)$, of which the general form is
\begin{align}
\mathrm{Prox}_{\Omega}\left(Z\right)=\arg\min_{\Gamma}\frac{1}{2}\Vert\Gamma-Z\Vert_{2}^{2}+\Omega\left(\Gamma\right). \label{eq:prox-1-1}
\end{align}
At each step, the sparse estimator $\widetilde{W}^l_t$ is the projection of $W^l_t$ onto the subspace of support set of $\Gamma^l_t$, i.e., Eq.~\eqref{eq:slbi-show-d} in which 
\begin{align}
\mathrm{Proj}_{\mathrm{supp}(\Gamma)}\left(W^l(i,j)\right) = \begin{cases} W^l(i,j), \ \ \mathrm{if} \ \Gamma^l(i,j) \neq 0 \\ 0, \ \ \ \mathrm{otherwise} \end{cases}
\end{align}
Starting from a null model with $\Gamma^l_0 = \widetilde{W}^l_0 = 0$, the $S^2$-LBI evolves into models $\{\Theta_t, W^l_t,\Gamma^l_t,\widetilde{W}^l_t\}$ with different sparsity levels, until the full model in which all parameters in $\Gamma^l$ are non-zeros, often-over-fitted. To avoid over-fitting, the early stopping regularization is often required. 

Under linear model ($\mathcal{L}$ is the squared loss function) implemented with gradient descent, Eq.~\eqref{eq:slbi-show} degenerates to Split LBI \cite{huang2016split} with $W$ representing a vector parameter, which is proposed to achieve the structural sparsity ($DW$ is sparse). It was proved that under linear model, the Split LBI can achieve model selection consistency, i.e. (i) the features selected at earlier steps belong to the true signal set $\mathcal{S} = \mathrm{supp}(DW^{\star})$, where $W^{\star}$ is the true signal (ii) successfully recover $\mathcal{S}$ at some iteration. Further, with $D = I$ and $\nu \to 0$, it degenerates to \emph{Linearized Bregman Iteration} (LBI), which is firstly proposed for image restoration with Total Variation penalty \cite{osher2005iterative}. Although $D = I$ in our scenario, we also adopted the variable splitting term with $\nu > 0$, since it was proved that such a term can improve the model selection consistency, especially with larger $\nu$ \cite{huang2016split}. The $S^2$-LBI can be regarded as the generalization of Split LBI, in terms of (1) general loss function (2) matrix parameters (3) mini-batch stochastic gradient descent. 

At a first glance at loss function $\bar{\mathcal{L}}$ in Eq.~\eqref{eq:slbi} and iterative scheme~\eqref{eq:slbi-show}, one may ask following two questions: (1) \textbf{\emph{The loss function  seems not including regularization function, then how $S^2$-LBI achieves sparsity?}} (2) \textbf{\emph{What's the loss function of Eq.~\eqref{eq:slbi-show} is optimizing? }}

In fact, different from the common iterative algorithm in which the estimator in the final step tries to converge to the minimizer of a certain loss function, the $S^2$-LBI is the discretization of the following differential inclusion with $\alpha \to 0$, 
\begin{subequations} \label{eq:slbi-iss-show} 
	\begin{align}
	\dot{\Theta}_{-W^l,t} & =-\nabla_{\Theta_{-W^l}}\bar{\mathcal{L}}\left(\Theta_t,\Gamma^l_{t}\right)\label{eq:slbi-iss-show-a}\\
	\frac{\dot{W}^l_{t}}{\kappa} & =-\nabla_{W^l}\bar{\mathcal{L}}\left(\Theta_t,\Gamma^l_{t}\right)\label{eq:slbi-iss-show-b}\\
	\dot{\Gamma}^l_t + \frac{\dot{\rho}^l_t}{\kappa} & =-\nabla_{\Gamma}\bar{\mathcal{L}}\left(W_{t},\Theta_{-W_t},\Gamma_{t}\right)\label{eq:slbi-iss-show-c}\\
	\rho^l_t & \in \partial  \Vert \Gamma^l_t \Vert_{1} \label{eq:slbi-iss-show-d}
	\end{align}
\end{subequations}
where $\rho^l_t$, as the subgradient of $\ell_1$ penalty, together with $\Theta_t, \Gamma^l_t$, are right continuously differentiable, $\dot{\rho}^l_t,\dot{\Theta}_t,\dot{\Gamma}^l_t$ denote right derivatives in $t$ of $\rho^l_t, \Theta_t, \Gamma_t$, and $\rho^l_0 = 0, \Gamma^l_0 = 0$, $\Theta_0$ is randomly initialized from Gaussian Distribution. We called such a differential inclusion as \emph{Stochastic Split Linearlized Bregman Inverse Scale Space} ($S^2$-LBISS) \cite{bregman}. The $t$ is regularization hyper-parameter, which determines the sparsity level and hence plays the similar role as $\frac{1}{\lambda}$ in Eq.~\eqref{eq:loss_function}. Compared with $\ell_{1}$ penalty which has to run optimization problem several times, a single run of LBI generates the whole regularization path, hence can largely save the computational cost. Here we give some implementation details of $S^2$-LBI (Eq~\eqref{eq:slbi-show}).

\begin{itemize}
	\item  The $\kappa$ is the damping factor, which enjoys the low bias with larger value, however, at the
	sacrifice of high computational cost. The $\alpha$ is the step size. In \cite{huang2016split}, it has
	been proved that the $\alpha$ is the inverse scale with $\kappa$ and should be small enough to ensure the
	statistical property. In our scenario, we set it to 0.01/$\kappa$.
	\item  The $\nu > 0$ controls difference between $\widetilde{W}^l$ and $W^l$, and it is the trade-off between model selection consistency and parameter estimator error. One one hand, as discussed earlier, larger $\nu$ can improve statistical property; on the other hand, too large value of $\nu$ can worsen the estimation of parameters. In our experiments, $\nu$ is often set with appropriately large value, which is also adopted in \cite{sun2017gsplit,zhao2018msplit}. Particularly, \cite{zhao2018msplit} discussed that features can be orthogonally decomposed into strong signals (mainly interpretable to the model), weak signals (with smaller magnitude compared with strong ones) and null features. It was proved that when $\nu$ is appropriately large, as the dense estimator $W$ can enjoy better prediction power by leveraging weak signals. 
	\item The $\Gamma$ at each step can be explicitly given by simplifying Eq.~\eqref{eq:prox-1-1},
	\begin{itemize}
		\item $\Gamma^{g} = \kappa \cdot \max(0,1 - 1/\Vert Z^{g} \Vert_2)Z^{g}$ for $g \in \{1,...,G\}$, when $W$ denotes the filter parameters, i.e. $W_{\mathrm{conv}}$.
		\item $\Gamma_{i,j} = \kappa \cdot \max(0,1 - 1/\Vert Z(i,j) \Vert_2)Z(i,j)$ for $i \in \{1,...,p_1\}$ and $j \in \{1,...,p_2\}$ when $W \in \mathbb{R}^{p_1 \times p_2}$ denotes the multiplication matrix before the fully connected layer, i.e. $W_{\mathrm{fc}}$.
	\end{itemize}
\end{itemize}

\subsection{From Boosting with Sparsity to $S^{2}$-LBI \label{subsec:Boosting-with-Sparsity}}

The $S^{2}$-LBI can be understood as a form of boosting as gradient
descent with structural sparsity via differential inclusion \cite{Huang2017Boosting}.
Particularly, the gradient descent step in~\eqref{eq:slbi-show-a}
and~\eqref{eq:slbi-show-b} can be regarded as a boosting procedure
with general loss \cite{Mason1999boosting}. For example, $L_{2}$-Boost
\cite{B2003Boosting} with square loss function can be firstly traced
back to Landweber Iteration in inverse problems \cite{Yuan2007On}.
To pursue sparsity, one can adopt FS$_{\epsilon}$-boosting which
updates parameters as follows:
\begin{align}
W^{d^{\star}}_{k+1}=W^{d^{\star}}_{k}-\epsilon\cdot\mathrm{sign}(g^{d^{\star}}_{k})\label{eq:update}
\end{align}
where $g_{k}$ is the gradient of $\bar{\mathcal{L}\left(\cdot\right)}$ w.r.t. $W_{k}$
and $d^{\star}:=\arg\max_{(i,j)}|g^{i,j}_{k}|$, and $\epsilon\in(0,1)$
is a small step size to avoid overfitting. The update~Eq~\eqref{eq:update}
can be characterized as a steepest coordinate descent, w.r.t., $\ell_{1}$-
norm: $\Delta=\arg\max\left\{ \langle v,g_{k}\rangle\mid\Vert v\Vert_{1}:=\sum_{i,j}\left|A_{i,j}\right|\right\} $.

\textbf{Mirror Descent}. We can achieve sparsity from a more
general perspective, \emph{i.e}. mirror descent algorithm (MDA) \cite{krichene2015accelerated}, which
is also called proximal gradient descent by extending the Euclidean
projection to proximal map associated with a sparsity-enforced Bregman
Distance. In our scenario, we consider dynamics 
\begin{subequations} \label{eq:mirror-iss-show}
	\begin{align}
	\dot{\Theta}_{t} & =-\nabla_{\Theta}\bar{\mathcal{L}}\left(W_t,\Theta_{-W_{t}},\Gamma_t\right)\label{eq:mirror-iss-show-a}\\
	\dot{Z}_{t} & =-\nabla_{\Gamma}\bar{\mathcal{L}}\left(W_t,\Theta_{-W_{t}},\Gamma_t\right)\label{eq:mirror-iss-show-b} \\
	\mathcal{B}^{t} & =\partial \Psi^{\star}(\mathcal{U}^{t})\label{eq:mirror-iss-show-c}
	\end{align}
\end{subequations}
where $\mathcal{B}\overset{\Delta}{=} [\Theta^{\top},\Gamma^{\top}]^{\top}$, $\mathcal{U} \overset{\Delta}{=} [\Theta^{\top},\Gamma^{\top}]^{\top}$. The $\Phi(\cdot)$ is a general regularization function that is often convex, and $\Phi^{\star}(z) = \max_{x} \langle x,z \rangle - \Phi(x)$. To pursue sparsity, we take $\Psi(\mathcal{B})=\Vert\Gamma\Vert_{1}+ \frac{1}{2\kappa} \left( \Vert W^l \Vert_{2}^2  +  \Vert \Gamma \Vert_{2}^2 \right) + \frac{1}{2} \Vert \Theta_{-W^l} \Vert_{2}^2$, then~Eq~\eqref{eq:mirror-iss-show} turns to $S^2$-LBISS (Eq.~\eqref{eq:slbi-iss-show}). 

Similarly, we can also write the discretized form of Eq.~\eqref{eq:mirror-iss-show}. It was proved in \cite{krichene2015accelerated} that with smooth $\Psi$ and as $t \to \infty$, we have
\begin{align}
D_{\Phi}(\mathcal{B}_t,\mathcal{B}^{\star}) = \Psi(\mathcal{B}_t) - \Psi(\mathcal{B}^{\star}) - \langle \partial \Phi(\mathcal{B}^{\star}), \mathcal{B}_t-\mathcal{B}^{\star} \rangle \nonumber 
\end{align}
where $\mathcal{B}^{\star} \overset{\Delta}{=} \arg\min_{\Theta,\Gamma}\bar{\mathcal{L}}(\Theta,\Gamma)$, which is however not satisfying the statistical property (e.g. model selection consistency) in the sparsity problem, since it is often over-fitted. Hence, to achieve sparsity, rather than the final step of $S^2$-LBI, we are more interested in the regularization solution path and the early stopping scheme that can avoid over-fitting.

\section{Forward/Backward Selection by $S^{2}$-LBI }
In this section, we apply $S^2$-LBI on expanding and pruning a neural network. The centered idea of these two tasks is by replacing SGD with $S^2$-LBI to train the network.  

\subsection{Forward Selection: Expanding a Network}

Since $S^2$-LBI evolves from null model to full model, and tend to select more parameters for $\Gamma$ as the algorithm iterates, we can apply it to expand a network. We take $W_{conv}$ in a certain layer (say, $l$-th layer)
as an example for the illustration of how to expand a network using $S^2$-LBI. Starting from
a small number of filters (e.g. 2) in convolutional layer of network,
more and more filters tend to be with non-zero value as the algorithm iterates.
At each step, we can compute the ratio of $\widetilde{W}^{g}$ to be non-zero. If this ratio passes a pre-set threshold, we add more convolutional filters (which are randomly initialized)
into $W_{k}$, and continue learning the network from training
data. More specifically, suppose at iteration $k$, the $l$-th layer has $G_k$ filters. With pre-defined selection criterion $s^{l}_{k}:= \#\{g \in [G_k]: \Vert \widetilde{W}_{conv,k}^{g} \Vert_2 > 0\} / G_k$, we determine whether $m$ new filters are added: If the selection criterion $s^{l}_{k}$ exceeds
a certain threshold $T$, then $m$ filters will be added, i.e., the number of filters turn to $G_k + m$. The newly added parameters are randomly
initialized, while the other parameters are kept the same. We call this process as Forward Selection
by $S^{2}$-LBI.

Note that the regularization parameter $t$ in Eq. \eqref{eq:slbi-iss-show} (or $t_{k}=k\cdot\alpha$ in $S^2$-LBI) is the trade-off between under-fitting and over-fitting. Therefore, unlike the optimization algorithm in which the estimator of the last step is the optimal one, the $S^{2}$-LBI can be
understood as returning a whole regularization solution path from under-fitting to over-fitting, in which
the estimators with earlier steps are often under-fitted, and then find the optimal point at some step, until finally over-fitted in the current parameter space as iterates. Hence, to avoid saturating to a local optimal in the current parameter space, we stop training when the algorithm meets the selection criterion, and lift to a higher parameter space by adding new parameters, in order to find ``better" optimal point.

\subsection{Backward Selection: Pruning a Network \label{subsec:Simplifying-a-network}}

On the other hand, the regularization solution
path $\{W_{k},\widetilde{W}_{k},\Gamma_{k}\}$ computed
by $S^{2}$-LBI can also be utilized to post-simplify a network when the training process is completed. Specifically, we propose
an \textbf{\emph{Order Strategy}} and \textbf{\emph{Magnitude Strategy
}} by setting some filters/features in $\widetilde{W}_{k}$ to 0. In more details, the \textbf{\emph{Order Strategy}} records 
the order that each filter/parameter in $\left\{ \widetilde{W}\right\} $ is selected along the solution
path. We use $\mathbf{E}$ to record the first epoch that a filter/feature to be non-zeros. The less number of $E$, the more importance of the filter/parameter. 
Besides, when the training process is completed, we can also compute the $L_{2}$ magnitude
$\mathbf{M}$ of each filter/parameter. The larger the value, the more importance of the filter/feature. We then measure the importance of filter/parameter
as follows, 
\begin{equation}
Sc=\lambda_{1}\cdot\mathbf{M}-\lambda_{2}\cdot\mathbf{E}\label{eq:score}
\end{equation}
where $\lambda_{1},\lambda_{2}>0$. The descending order of
$Sc$ ranks the importance of filters/parameters in the network. We can
directly set the filters/paramters in $W$ with low rank in $Sc$ as 0. In that follows, we give some explanation of why we use $\mathbf{E}$ and $\mathbf{M}$ to measure the importance. 

\textbf{Explanation of $\mathbf{E}$}. As mentioned earlier,
the regularization solution path $\{\widetilde{W}_{k},\Gamma_{k}\}$
is proved \cite{Splitlbi} in Genearlized Linear Model (GLM) to enjoy the model selection consistency
that the features selected before some point, belong to the true signal
set. That means, the \textbf{\emph{Order Strategy}} is reasonable
to measure the importance of the selected features. Specifically,
the filter selected at smaller $k$ may be more correlated with $y$. After $k$ passes a threshold, the filters selected
are mostly fit random noise, which should hence be removed.

\textbf{Explanation of $\mathbf{M}$}. Zhao \emph{et al.} \cite{zhao2018msplit} orthogonally
decomposed all features into strong signal, weak signal and null features. The ``strong" signals refer to the features
selected via sparsity regularization, which corresponds to $\widetilde{W}$
and accounts for the main component of the model hence often have
larger magnitudes than rest of the features. The weak signals correspond
to the elements with smaller magnitude than strong signals, but larger magnitude than null features. By ranking the 
ranking the magnitude of learned parameters except for strong signals,
the weak signals can be detected unbiasedly as long as $\nu$ is large enough. Moreover, it can be proved theoretically and experimentally that equipped with
such weak signals, the dense estimator ($W$) outperforms sparse estimator
($\widetilde{W}$) in the prediction. Thus, the \emph{\textbf{Magnitude Strategy}} can be leveraged to differentiate weak signals from random noise.

%\scriptsize{ \begin{algorithm}[h!]   
%\caption{$S^2$-LBI on multiple layers}   
%\label{alg:lbi-m}    	
%\begin{algorithmic}[1]          
%\STATE  $\textbf{Input:}$ Learning rate $\eta$, $\nu > 0$, step size of LBI $\alpha$,  damping factor $\kappa > 0$, $\mathcal{X}$ and $y$          	
%\STATE $\textbf{Initialize:}$ $k=0$, $\Theta^{k}$ is initialized randomly, $\Gamma_1^k = Z_1^k= 0$, ..., $\Gamma_L^k = Z_L^k= 0$. Denote $[W^k] = \{ W_1^{k},..., W_L^{k} \}$, $[\Gamma^k] = \{ \Gamma_1^{k},..., \Gamma_L^{k} \}$ \STATE $\textbf{Iteration}$           
%\STATE \quad    $\Theta_{-[W^k]}^{k+1} = \Theta_{-[W^k]} - \eta \nabla_{\Theta_{-[W^k]}} L(\Theta_{-[W^k]}, [W^k], [\Gamma^k] )$\newline          \# \emph{LBI update at L layers}       
%\STATE \quad  \textbf{For} \bm{$l = 1,...,L$}          
%\STATE \quad \quad \bm{  $W_l^{k+1} = W_l^{k} - \kappa \alpha \nabla_{W}L(\Theta_{-[W^k]}, [W^k],[\Gamma^k]  )  }$                   
%\STATE \quad \quad \bm{  $Z_l^{k+1} = Z_l^{k} - \alpha \nabla_{\Gamma} L(\Theta_{-[W^k]}, [W^k],[\Gamma^k]  ) } $                     \STATE \quad \quad \bm{  $\Gamma_l^{k+1} = \kappa \mathrm{Prox}_{J}(Z_l^{k+1})   }$ 	  
%\STATE \quad \quad  \bm{  $\widetilde{W}_l^{k+1} = W_l^k \circ \left[ 1 \{(i,j) \in S_l^{k+1}\} \right] }$  
%\STATE \quad \textbf{End}     
%\STATE $\textbf{Output:}$ $\{\Theta_{-[W^k]}, [W^k], [\widetilde{W}^k], [\Gamma^k] \}$  	
%\end{algorithmic}  \end{algorithm} } \normalsize 

\section{Related works}

\noindent \textbf{Network Regularization.} It is essential to regularize
the networks, such as dropout \cite{srivastava2014dropout} preventing
the co-adaptation, and adding $L_{2}$ or $L_{1}$ regularization
to weights. In particular, $L_{1}$ regularization enforces the sparsity
on the weights and results in a compact, memory-efficient network
with slightly sacrificing the prediction performance \cite{l1_memory}.
Further, group sparsity regularization \cite{yuan2006model} can also
been applied to deep networks with desirable properties. Alvarez \emph{et
	al.} \cite{group_spars_network} utilized a group sparsity regularizer
to automatically decide the optimal number of neuron groups. The structured
sparsity \cite{l12_norm,yoon2017combined} has also been investigated
to exert good data locality and group sparsity. In contrast, the regularization
term of $S^{2}$-LBI proposed can not only enforce the structured
sparsity, but also can efficiently compute the regularization solution
paths of each variable.

\noindent \textbf{Expanding a Network.} Most previous works of expanding
network are mostly focused on transfer learning/ life-long learning
\cite{Wang2017Growing,Tom1995lifelong,lifelong_iid,zhizhong2016eccv},
or knowledge distill \cite{hinton_distill}. In contrast, to the best
of knowledge, this work, for the first time, can enlarge a network
in a more general setting in the training process: our strategy is
more general and can start from a very small network. The inverse
process of growing network, is to simplify a network. Comparing to
network pruning algorithms \cite{zhu2016trained,zhou2017incremental,jaderberg2014speeding,zhang2016accelerating,han2015learning,li2016pruning,greedy_filter_pruning,soft_filtering},
our strategy directly removes less important parameters by using the
solution path computed by $S^{2}$-LBI. The backward selection algorithm
does not introduce any additional computational cost in learning to
prune the networks.

\noindent \textbf{regularization solution path.} Regularization path
is a powerful statistic tool to analysis data and statistic model,
it can be used for feature selection. Usually, in linear model, we
could set different regularization parameter, and get the regularization
path by fitting the model for several times. In contrast, $S^{2}$-LBI
is a novel algorithm that could generate regularization solution path
in one training procedure. Our $S^{2}$-LBI does not only provide
an efficient way of training deep neural network, but also give an
intrinsic mechanism of analyzing deep network. Thus built upon $S^{2}$-LBI,
we can expand a network. In GLM, our $S^{2}$-LBI will be degenerated
into Split LBI \cite{Splitlbi}, which satisfies path consistency property that features belong to the true signal set will be firstly selected. 

\section{Experiments}

We conduct  experiments to validate  $S^{2}$-LBI on MNIST and
CIFAR-10. MNIST is a widely used dataset with 60,000 gray-scale
handwritten digits of 10 classes. CIFAR-10 is a well-known object
recognition dataset, which contains of 60,000 RGB images with size
$32\times32$. It have 10 class with 6000 images per class. We use
the standard supervised training and testing splits on both datasets.
20\% of training data are saved as the validation set. The classification
accuracy is reported on test set of each dataset. Source codes/models
would be download-able.

\subsection{Expanding a Network}

Generally, increasing the number of neurons in the FC layer have
much less importance than adding the filters of convolutional layer.
For example, Han \emph{et al.} \cite{han2015learning} can directly
prune the fully connected layer, from dense into sparse, without sacrificing
the performance. Thus our experiments are designed to show that our
$S^{2}$-LBI can successfully and efficiently increase the number
of convolutional filers in both dataset. The resulting network can
have comparable performance with much less parameters. 
\begin{table}[ht]
	\centering
	{\scriptsize{}}%
	\begin{tabular}{c|c|c|c}
		\hline 
		\multirow{2}{*}{} & \multicolumn{1}{c|}{{\small{}{}MNIST}} & \multicolumn{2}{c}{{\small{}{}CIFAR-10}}\tabularnewline
		\cline{2-4} 
		& {\small{}{}Acc. } & {\small{}{}Acc.(1L) } & {\small{}{}Acc. (2L)}\tabularnewline
		\hline 
		\hline 
		{\small{}$S^{2}$-LBI (whole path)} & {\small{}98.30} & {\small{}63.56} & {\small{}74.59}\tabularnewline
		\hline 
		{\small{}{}SGD } & {\small{}{}98.17 } & {\small{}{}65.30 } & {\small{}{}74.51}\tabularnewline
		\hline 
		{\small{}{}Adam } & {\small{}{}98.16 } & {\small{}{}64.73 } & {\small{}{}75.75}\tabularnewline
		\hline 
		\hline 
		{\small{}{}$S^{2}$-LBI (forward selection) } & {\small{}{}98.40 } & {\small{}{}63.44 } & {\small{}{}74.23}\tabularnewline
		\hline 
	\end{tabular}{\small\par}
	\par
	\caption{\label{tab:The-results-of-training}The results of using different
		training algorithms to learn the expanded structure of $S^{2}$-LBI.
		``1L''/ ``2L'' indicate 1 / 2 convolutional layers.}
\end{table}

\subsubsection{Expansion on MNIST}

We expand a simple CNN network, which only contains one convolutional
and one fully connected layer. The initial network structure is defined
as conv.c1 ($1@5\times5$) -- Maxpooling($2\times2$) -- FC (144
neurons) -- 10 classes, where $1@5\times5$ means 1 filters with
size $5\times5$, the default activation function -- ReLU and and
the softmax in classification layer. In our experiments, we set the
threshold $T=80\%$. That means, if more than $80\%$ convolutional
filters in $W_{conv}^{k}$ are selected into $\widetilde{W}_{conv}^{k}$,
our $S^{2}$-LBI will automatically add 2 more convolutional filters
into this layer. The weighting connection of Fully Connected (FC)
layer will also be correspondingly adapted in connecting to the newly
added filters. The newly added parameters are initialized as \cite{he2015delving}.
\begin{figure}[ht]
	\centering{}\includegraphics[scale=0.6]{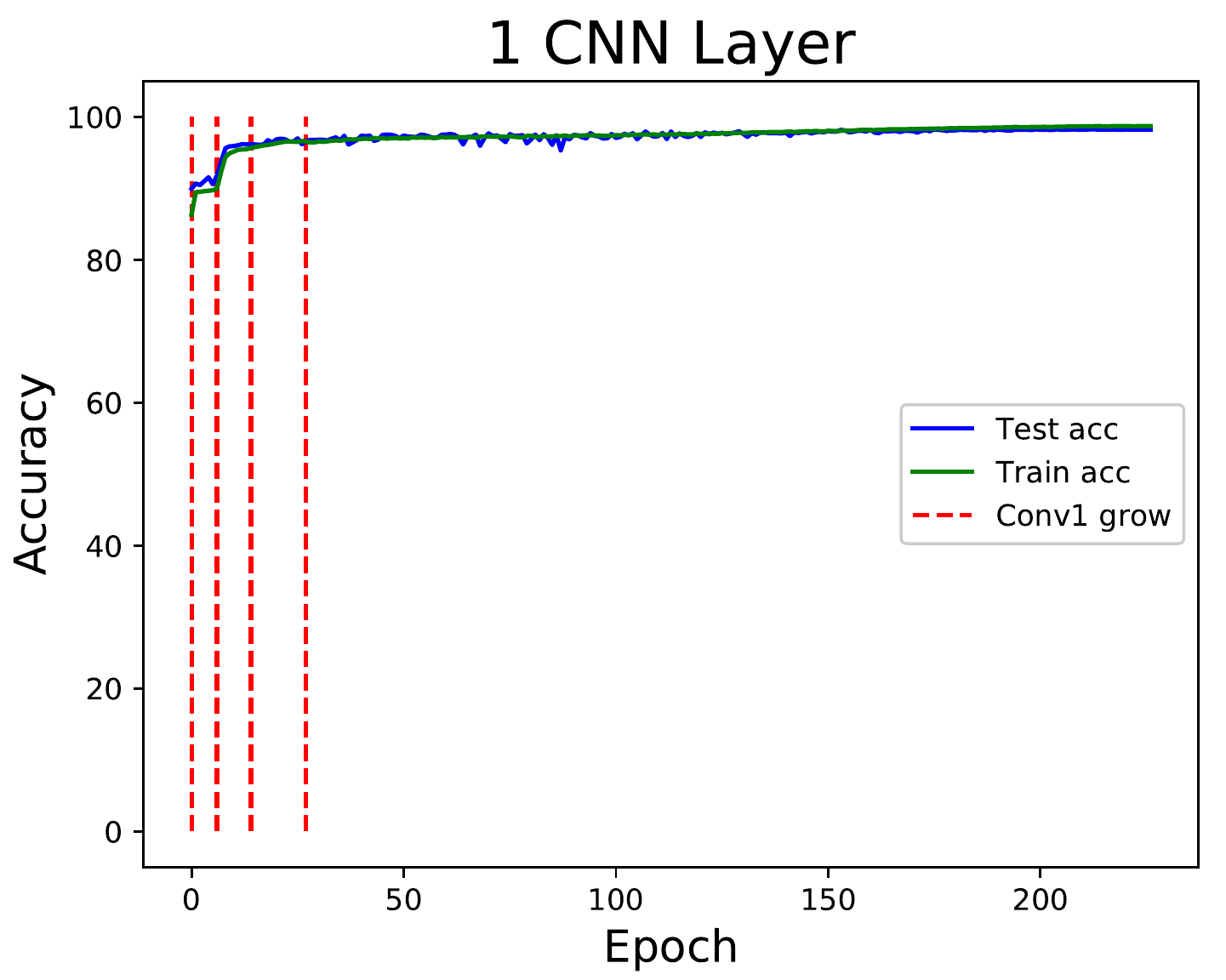}\caption{Results of expanding a network on MNIST dataset. The X-axis and Y-axis
		indicates the training epochs, and classification accuracy on the
		testing data. The red dash lines indicate 2 new filters are automatically
		added in that epoch by our $S^2$-LBI algorithm. Note that the training
		process can be early stopping at around 50 epochs; and we show the
		results of much more epochs to better understand the performance of
		our $S^{2}$-LBI. \label{mnist_growth} }
\end{figure}
\begin{figure}
	\centering{}%
	\begin{tabular}{c}
		\hspace{-0.2in}\includegraphics[scale=0.5]{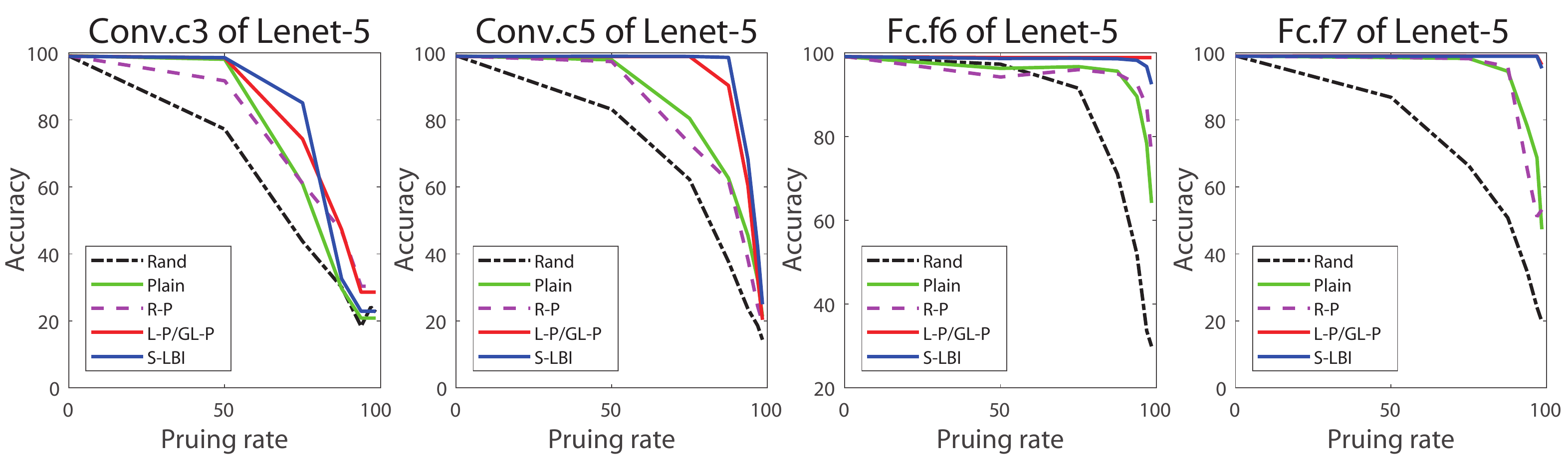}\tabularnewline
	\end{tabular}\caption{We simplify the LeNet-5 structure on (a) conv.c3 layer, (b) conv.c5
		layer, (c) fc.c6 layer, (d) fc.f7 layer. Several different competitors
		are compared. $S^2$-LBI algorithm is employed to train the network.\label{fig:We-simplify-mnist}}
\end{figure}

We show the results of expanding the network in Fig. \ref{mnist_growth}.
We can observe that, with only 7 filters (175 parameters) and 1 FC
layer (1440 parameters), the expanded network can achieve the classification
accuracy of $98.40\%$. The total parameters are 1.6K parameters.
In contrast, the LeNet-5 \cite{LeCunCNN} is pre-defined network structure
which is composed of 3 convolutional layers and 2 fully connected
layers. The classification accuracy of LeNet-5 is $\sim99.0\%$\cite{LeCunCNN}
with the parameters of 61.5K parameters. We use SGD, rather than $S^{2}$-LBI
algorithm to train the same boosting network; and the results are
$98.55\%$. This implies that our enlarged network can achieve comparable
results as LeNet-5 with much less parameters. More importantly, most
of parameters of our enlarged network is automatically added according
to the training data without human interaction.
\begin{figure}
	\begin{centering}
		\begin{tabular}{cc}
			\includegraphics[scale=0.5]{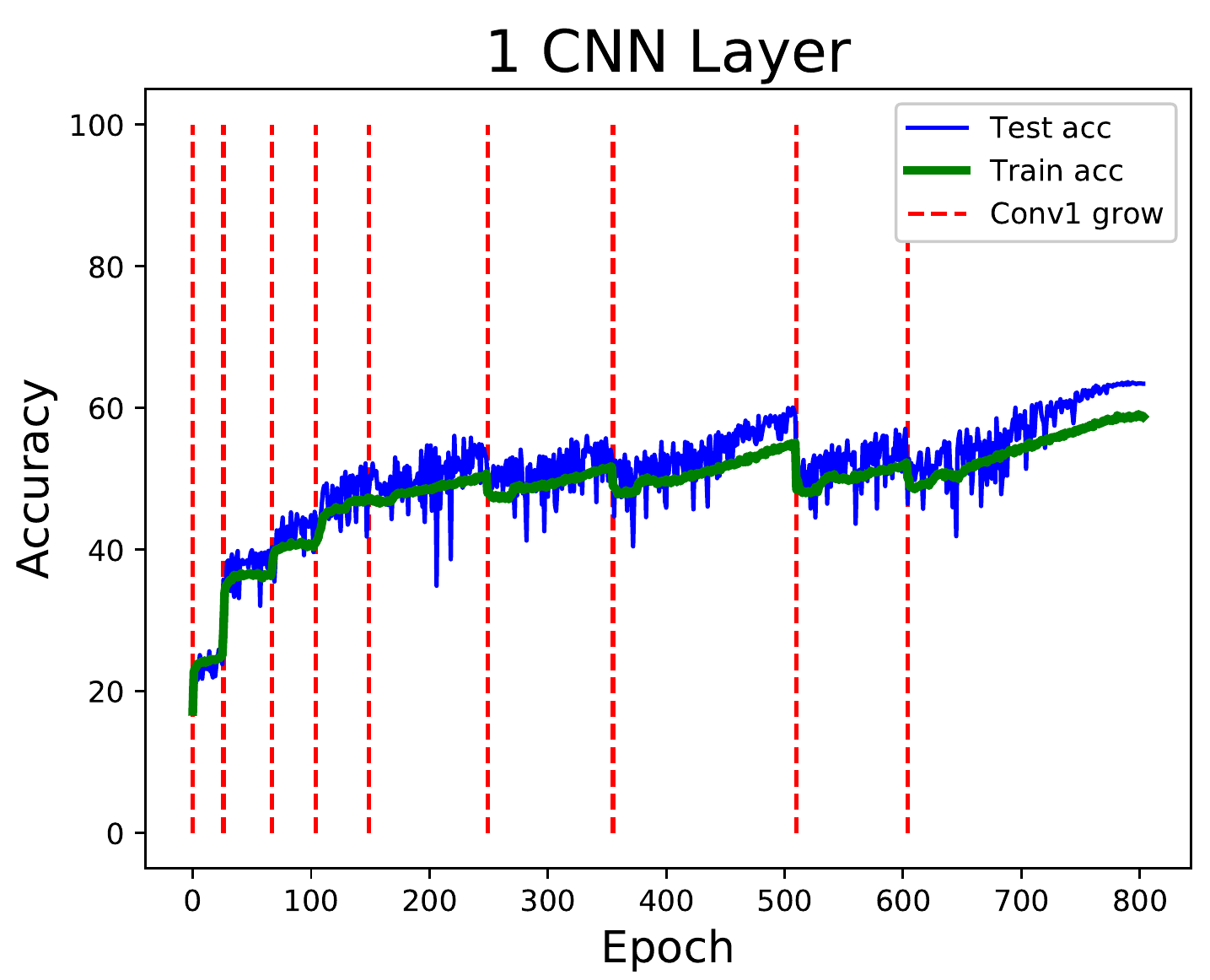}  & \includegraphics[scale=0.5]{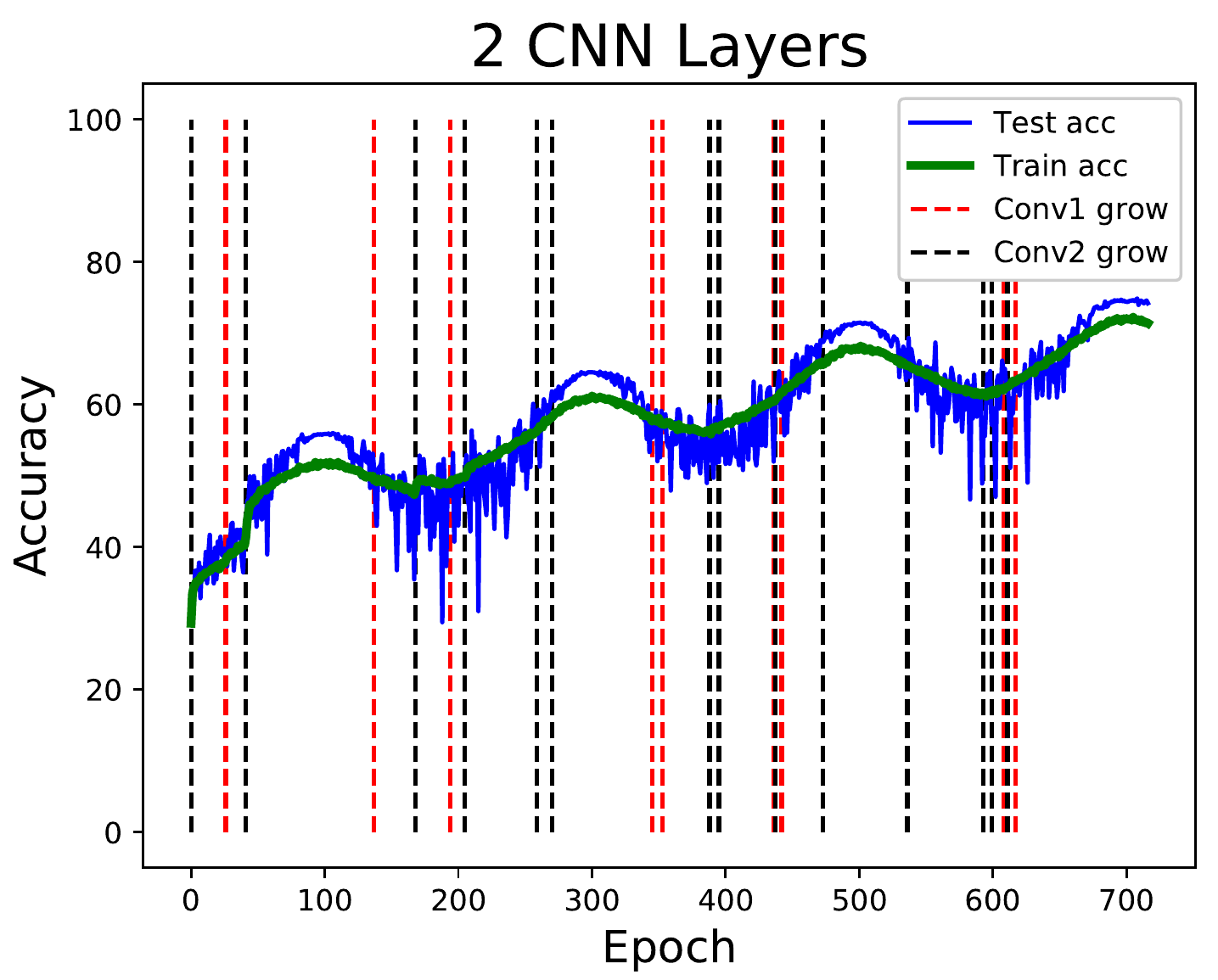}\tabularnewline
		\end{tabular}
		\par\end{centering}
	\caption{\label{fig:Booting-a-cifar10}Booting a network in CIFAR-10 dataset.}
\end{figure}

\begin{figure}
	\begin{centering}
		\includegraphics[scale=0.8]{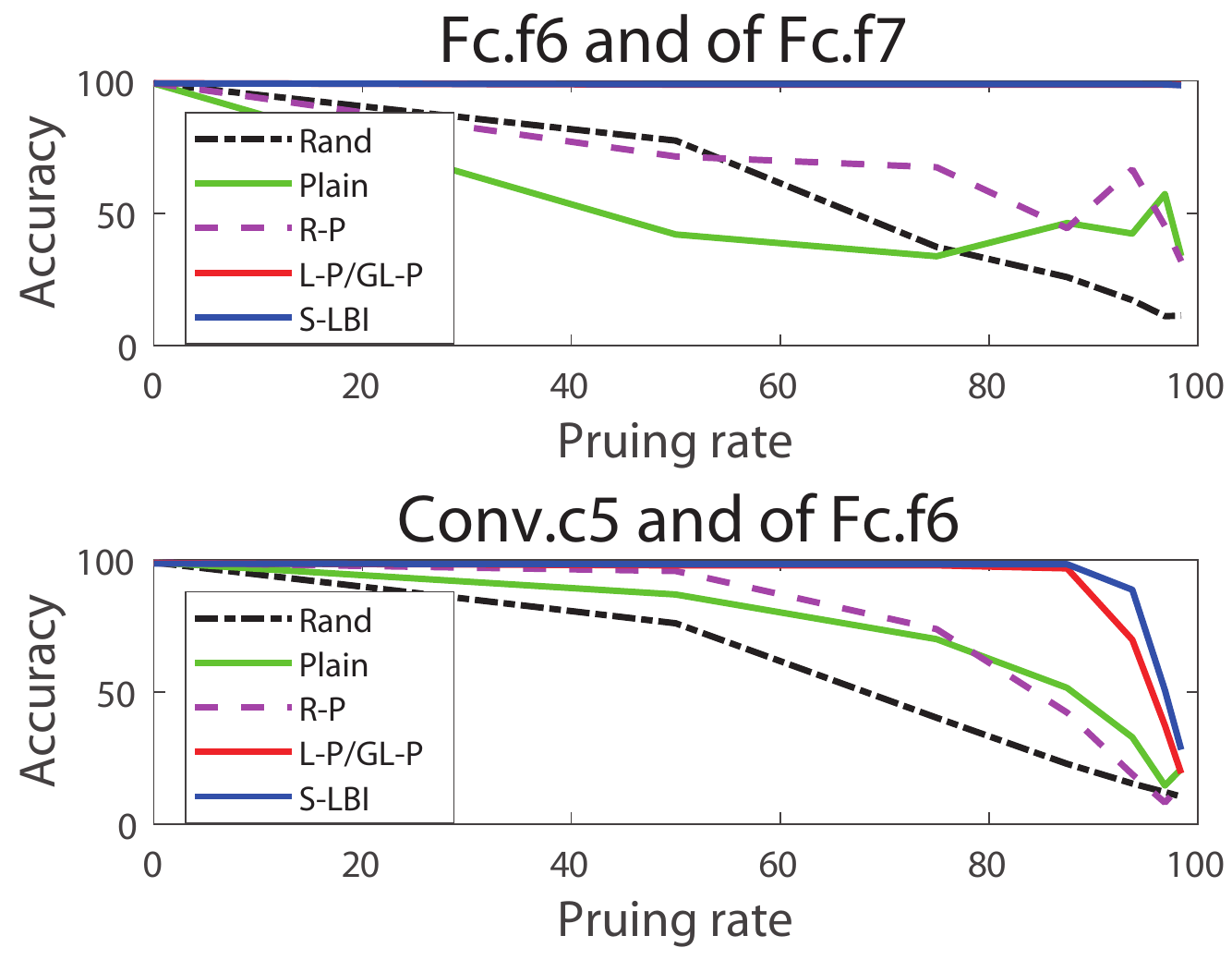} 
		\par\end{centering}
	\caption{\label{tab:Pruning-two-layers}Simplifying two layers in LeNet-5 on
		MNIST dataset. Each column indicates the percentage of parameter saved
		on these two layers. \emph{Com-Rat}, is short for the compression
		ratio of the total network, \emph{i.e.}, the ratio of saved parameters
		divided the total number of parameters of LeNet-5. }
\end{figure}

\subsubsection{Expanding a Network in CIFAR-10 }

We enlarge the CNN network in CIFAR-10 dataset. We adopt two initial
networks, (1) Network-1: conv.c1 ($1@5\times5$) -- Maxpooling($2\times2$)
-- FC (144 neurons) -- 10 classes; (2) Network-2: conv.c1 ($1@5\times5$)
-- Maxpooling ($2\times2$) --conv.c1 ($1@3\times3$) -- Maxpooling($2\times2$)
-- FC (36 neurons) -- 10 classes. We have default activation function
ReLU, and the softmax is used in classification layer. We have the
threshold $T=80\%$; and if the ratio of filters selected into $\widetilde{W}_{conv}^{k}$
is higher than the threshold, 2 more filters are automatically added
by $S^2$-LBI. The connections will be automatically added between FC layers.
Only the newly added parameters, which are initialized as \cite{he2015delving};
and we donot re-initialize the parameters of the other parts.

The results are shown in Fig. \ref{fig:Booting-a-cifar10}. The final
classification accuracy of Network-1 and Network-2 is $63.44\%$
and $74.23\%$ individually. In Network-1, our Forward Selection algorithm
results 17 convolutional filters; and Network-2 has 21 and 29 convolutional
filters. These results are comparable to those of state-of-the-arts
of 1 or 2 layer networks, e.g., 3-Way Factored Restricted Boltzmann
Machine (RBM) (3 layers)\cite{Hinton2010cvpr} : 65.3\%; (2) Mean-covariance
RBM (3 layers) \cite{Ranzato2010}: 71.0\%; (3) SGD: 65.30/74.51,
and (4) Adam : 64.73/75.75. 

To further compare these results, the resulting networks
by forward selection algorithm, are trained by different other optimization
algorithms, e.g., SGD, Adam, and $S^{2}$-LBI (whole path). Note that,
$S^{2}$-LBI indicates that we directly use the $S^{2}$-LBI to train
the fixed networks, rather than employing the Forward Selection algorithm
to gradually add filters. All algorithms are trained until it gets
converged. The results are summarized in Tab. \ref{tab:The-results-of-training}.
It shows that the results of our forward selection algorithm are comparable
to $S^{2}$-LBI (whole path). Note that both $S^{2}$-LBI (whole path)
and $S^{2}$-LBI (forward selection) gets the converged at the same
number of training epochs; but the forward selection algorithm has
much less computational cost, since the filters are gradually added.
Admittedly, the results of SGD, and Adam have slightly better results
than those of $S^{2}$-LBI. This is also reasonable. Our $S^{2}$-LBI
enforces the structure sparsity to the network, and it can be taken
as a set of feature selection algorithms. This is the main advantage
over the SGD, and Adam. The other weak features learned in SGD and
Adam, (but not selected by $S^{2}$-LBI ) may also be useful in the
prediction \cite{zhao2018msplit}. For example, Lasso does not necessarily
have better prediction ability than Ridge Regression. To sum up, the
results shows that our $S^2$-LBI algorithm can efficiently grow a network.
\begin{figure}
	\begin{centering}
		\begin{tabular}{c}
		\includegraphics[scale=0.45]{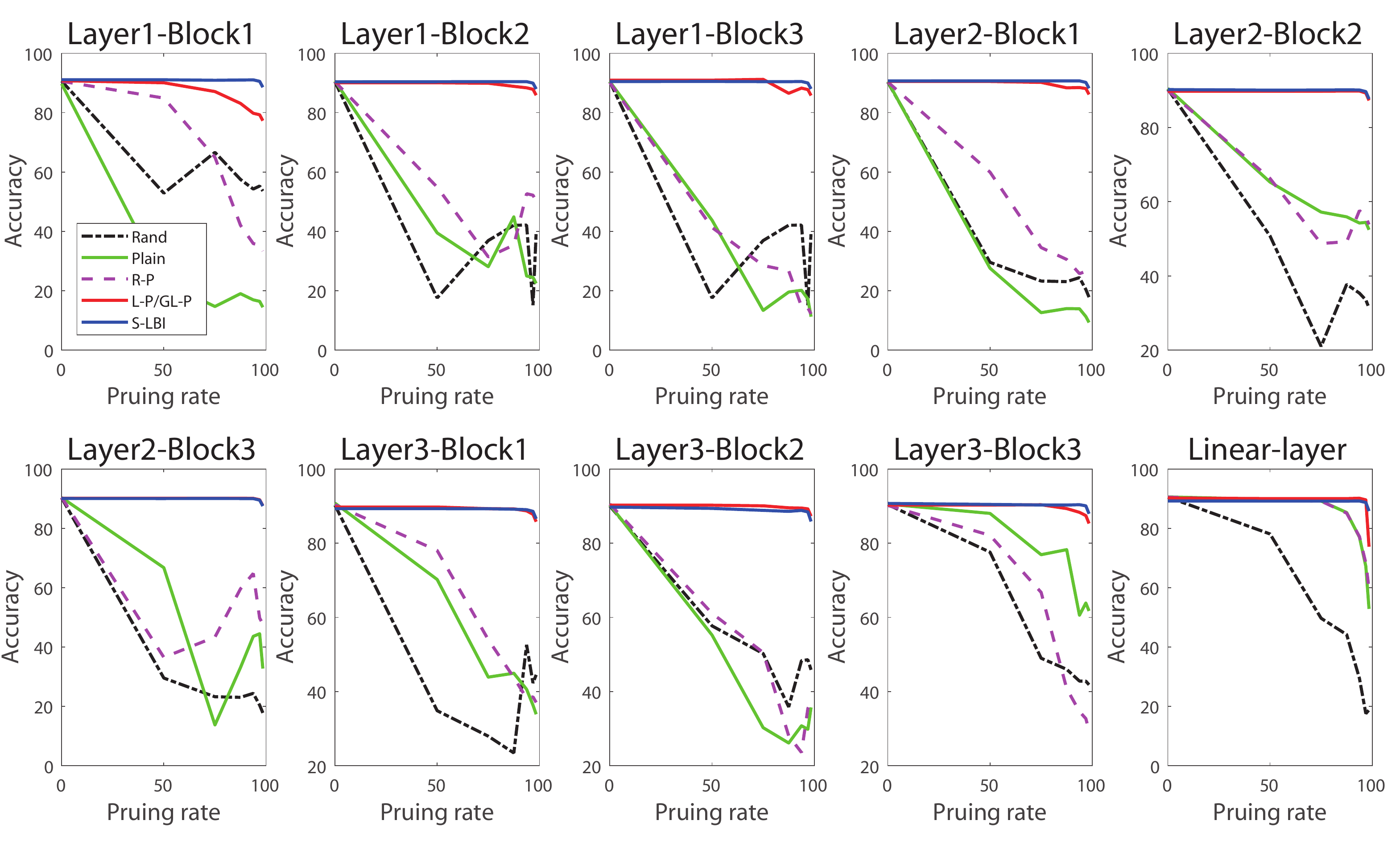}\tabularnewline
		\end{tabular}
		\par\end{centering}
	\centering{}\caption{We simplify the ResNet-20 structure trained on CIFAR-10 dataset. Several
		different competitors are compared. Please refer to supplementary
		for large figure.\label{cifar-10-simplifying} }
\end{figure}

\subsection{Simplifying networks}

\subsubsection{Simplifying networks on MNIST}

\begin{figure*}
	\centering{}\includegraphics[scale=0.4]{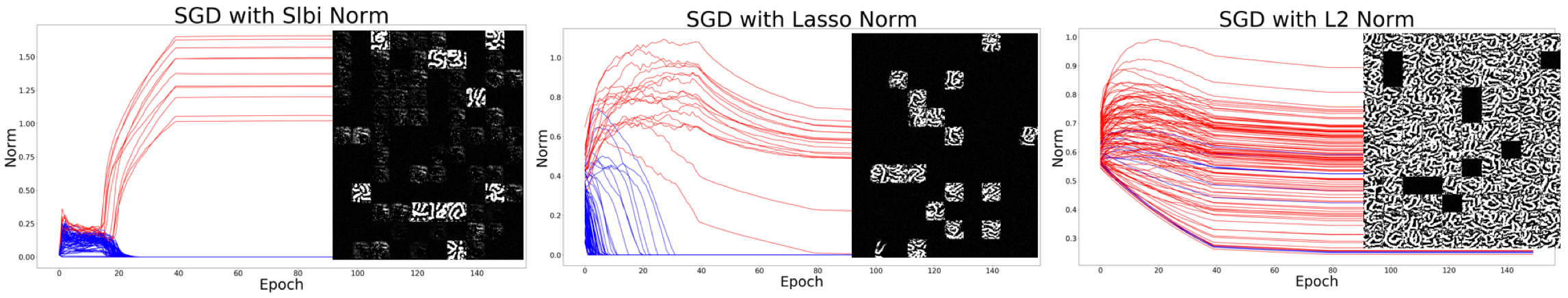}\caption{Visualization of solution path and filter patterns.
		X-axis and Y-axis are the training epochs, and filter magnitude 
		respectively.}
	\label{path_visualization} 
\end{figure*}

\textbf{Competitors}. We compare several other methods of simplifying
the networks. (1) \emph{Plain}: we train a LeNet-5 network; we use
$L_{2}-$norm on all parameters with the normalizing coefficient $5e-4$.
We rank the importance of weights/filters in term of their $L_{2}$
magnitude values in the descending order and remove those with low rank. 
This is a naive baseline of our simplifying network algorithm. (2)
\emph{Rand}. We randomly remove the weights or filters in the networks.
This is another naive baseline. (3) \emph{Ridge-Penalty }(\emph{R-P})
\cite{han2015learning}: we use $L_{2}-$norm to penalize one layer
of network. Different from Plain, the normalizing coefficient is set
as $1e-3$, which is cross validated by searching from $1e-1$ to
$1e-5$ on Lenet-5. (4)\emph{ Lasso-type penalty} or \emph{Group-Lasso-type
	penalty} (\emph{L-P / GL-P}): the L-P is used to prune the weights
of fully connected layers, and we employ the GL-P to directly remove
the filters of convolutional layers. We also cross validate the normalizing
coefficient, which is finally taken as $1e-4$. Note that all the
results are trained for one time, and we do not have fine-tuning step
after the network simplification.

\textbf{Simplifying one layer}. The results are shown in
Fig.~\ref{fig:We-simplify-mnist}. We use our $S^{2}$-LBI algorithm
to train the LeNet-5, and simplify each individual layer of LeNet-5,
while we keep the parameters of the other layers unchanged. Note that,
the simplified network is not fine-tuned by the training data again.
As compared in the Fig. \ref{fig:We-simplify-mnist}, we have the
following observations: (1) On two fully connected layers (fc.f6 and
fc.f7), both the L-P /GL-P and our simplifying network algorithm work
very well. For example, on the fc.f7 layer, our s only has 1.57\%
of the parameters on these layers. Surprisingly, our performance is
only 0.03\% lower than that of the original network. In contrast,
we compare such results with the baselines: \emph{Plain}, \emph{Rand},
and \emph{R-P}. There is significant performance dropping with the
more parameters removed. This shows the efficacy of our algorithm.
(2) On the convolutional layer (conv.c5), our results still also achieve
remarkable results. The conv.c5 layer has $48k$ out of $61k$ number
of parameters of Lenet-5. We show that ours saves 12.5\% of total
parameters of this layer (\emph{i.e.}, $42k$ number of parameters
can be removed on this layer) and the results get only dropped by
0.3\%. This demonstrates that our $S^{2}$-LBI indeed can select important
weights and filters from $\left\{ W^{k}\right\} $ into $\left\{ \widetilde{W}^{k}\right\} $
in the $k$-th training epoch. (3) The conv.c3 layer is another convolutional
layer in LeNet-5. We found that this layer is very important to maintain
a good performance of overall network. Nevertheless, our results are
still better than the other competitors.

\textbf{Simplifying more layers. }With the trained model
by our $S^{2}$-LBI, we can consider simplifying two layers together,\emph{
	i.e.}, \emph{fc.f6 + fc.f7}, or \emph{conv.c5+fc.f6 }layers, since
\emph{conv.c5} and \emph{fc.f6} have $48k$ and $10k$ number of parameters
out of the total $61k$ parameters in LeNet-5. The results are reported
in Fig. \ref{tab:Pruning-two-layers}. We can show that our framework
can still efficiently compress the network while preserve significant
performance. Furthermore, when we simplify the \emph{conv.c5} and
\emph{fc.f6} layers, our model can achieve the best and efficient
performance. With only 17.60\% parameter size of original LeNet-5,
our model achieves the performance as high as 98.47\%. Remarkably,
our simplifying algorithm does not need any fine-tuning and
re-training steps. This shows the efficacy of our
our $S^{2}$-LBI can indeed discover the important weights and filters
by using the solution path. Our best models will be downloaded online.

\subsubsection{Simplifying networks on CIFAR-10 }

We use  ResNet-20 \cite{he2016deep} as backbone network to conduct the experiments on CIFAR-10. We still compare
the competitors, including R-P, L-P / GL-P, and Rand. The hyper-parameters
of all methods are cross-validated.

\textbf{Simplifying one layer}. The results are compared
in Fig. \ref{cifar-10-simplifying}(a). We employ our $S^{2}$-LBI
algorithm to train the ResNet-20 structure. We simplify each individual
layer as the score in Eq \ref{eq:score}, while keep the parameters
of the other layers unchanged. The results show that our simplified
network can work better than the other competitors. This validates
the efficacy of our proposed $S^{2}$-LBI algorithm in training and
select the important filters and weights of network.

\begin{figure}
	\centering{}%
	\begin{tabular}{cc}
	\includegraphics[width=0.5\columnwidth]{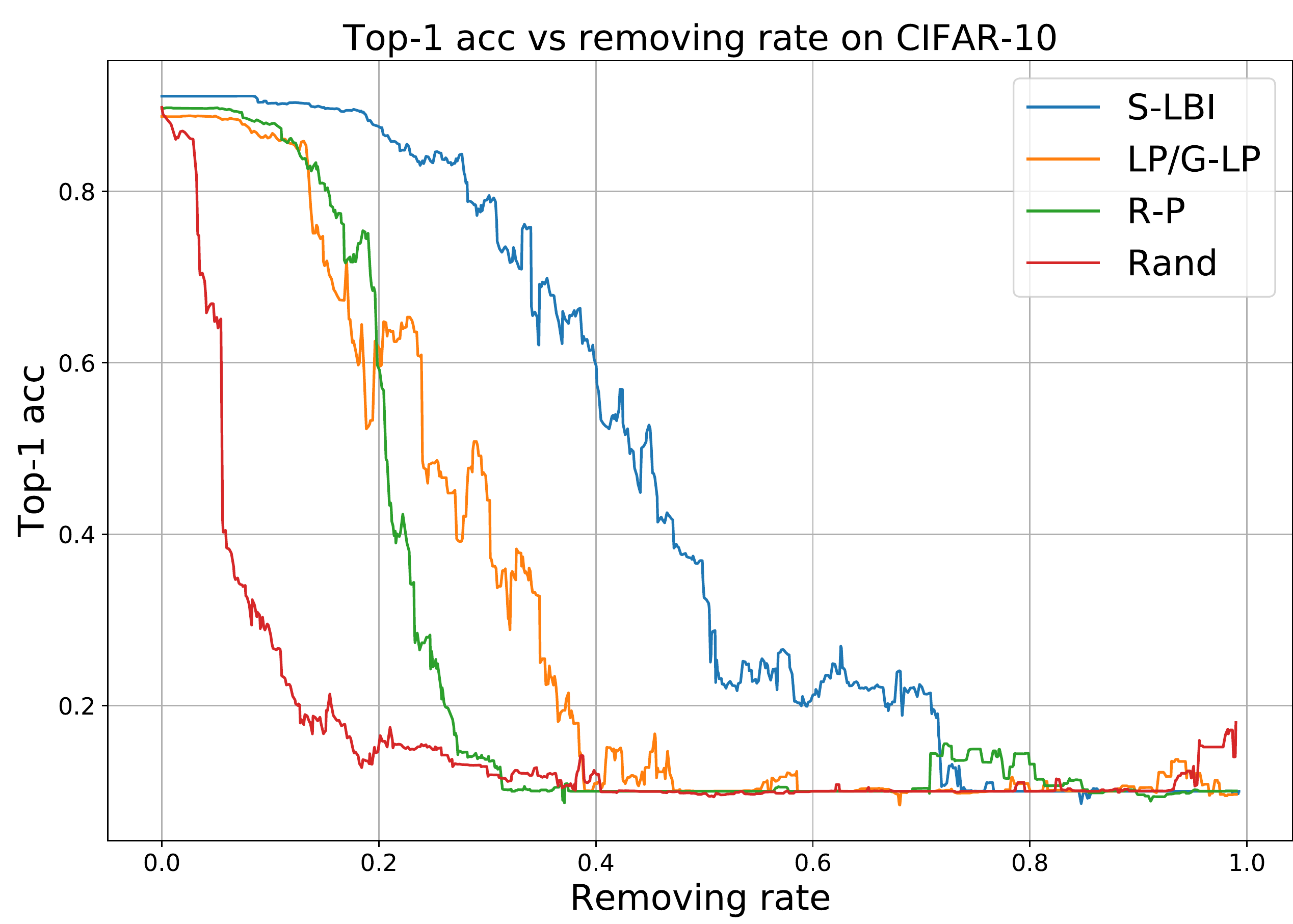}  & \includegraphics[width=0.5\columnwidth]{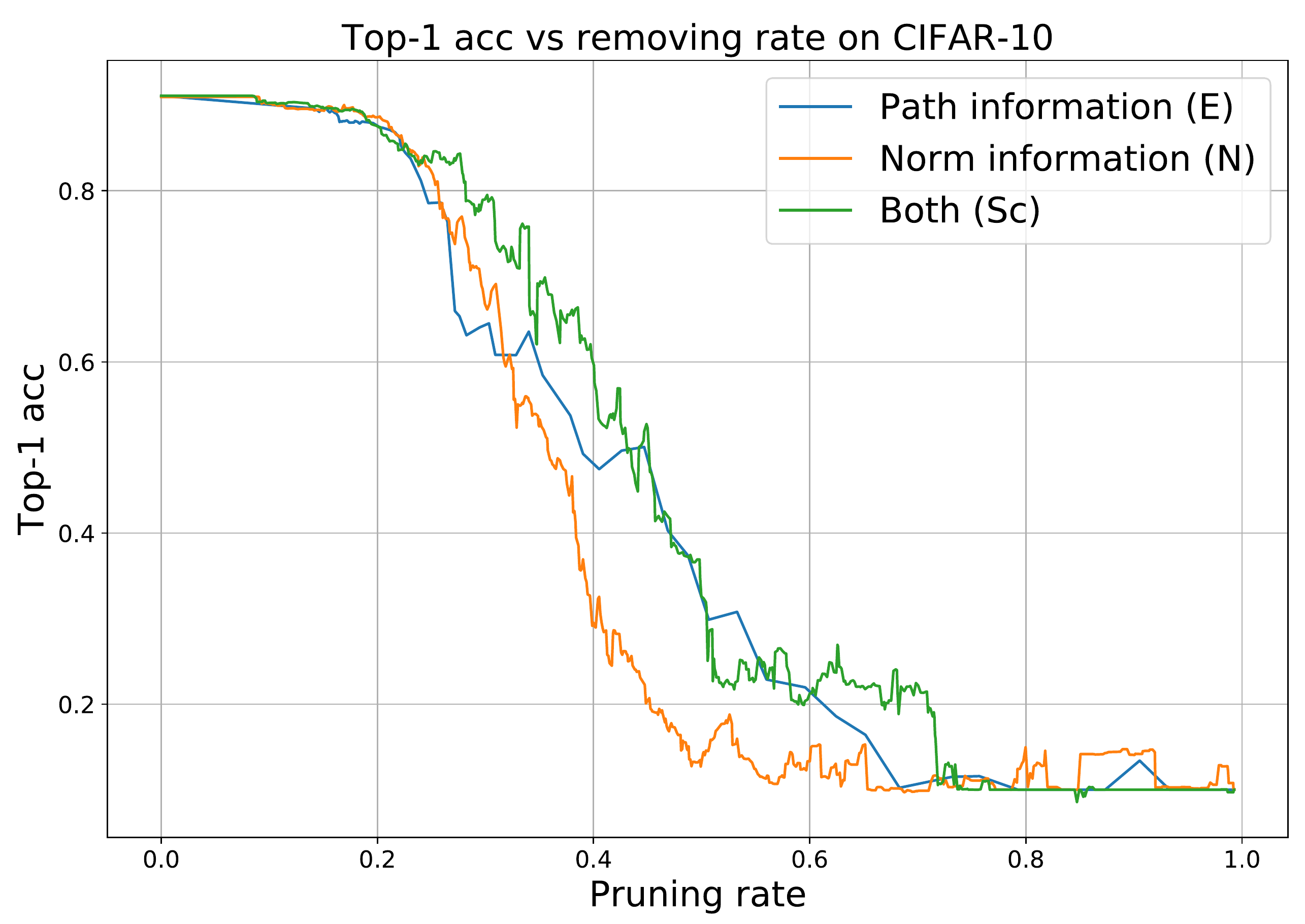}\tabularnewline
		(a) Simplifying  & (b) Different scores\tabularnewline
	\end{tabular}\caption{(a) Comparison of simplifying ResNet-20 in CIFAR-10. X-axis and Y-axis
		represent the removing rate, and top-1 accuracy respectively. We can
		see that our $S^{2}$-LBI algorithm could beat all of our baseline
		models. \label{cifar10_all} (b) Pruning performance, if we only consider
		solution path (i.e.. $E$) or we only consider the magnitude of filters
		(i.e., $N$). The results are from the experiment in reducing all
		layers of Resnet-20 on CIFAR-10 dataset. \label{path_vs_norm-1} }
\end{figure}

\textbf{Simplifying more layers. }We further conduct
experiments of simplifying more layers of ResNet-20. Particularly,
we train the network by $S^2$-LBI, and rank the importance of filters
by Eq \ref{eq:score}. The results are shown in Fig. \ref{cifar10_all}(a).
It shows that our simplified network has much better results than
others; this again, demonstrates the efficacy of $S^2$-LBI in efficiently and selectively training the network.
Note that, the ResNet-20 is relatively a compact structure of learning
data in CIFAR-10; thus with the pruning rate beyond $20\%$, all methods
get degraded dramatically.

\subsection{Ablation Study}

\textbf{Simplifying criterion.} The score of filters are weighted
by the vector $E$ of first selecting one filter, and magnitude
value vector $M$ of each filter at the converged epoch, as Eq \ref{eq:score}.
On CIFAR-10, we use the ResNet-20 as the backbone, and simplifying
the network by using $E$, $M$ or $Sc$ (Eq \ref{eq:score}).
The results are shown in Fig. \ref{path_vs_norm-1}. We remove the
filters one by one. Clearly, it shows that by using either the path
information ($E$), or the magnitude information ($M$), the network can
still be efficiently simplified, again thanks to our $S^{2}$-LBI
algorithm. Importantly, by combing both information, our algorithm
can achieve better performance.

\textbf{Visualization of Filters in Solution Path.} We visualize the
filters as the visualization algorithm \cite{Erhan2009Visualizing}
in order to show the filters learned by our $S^{2}$-LBI algorithm.
On MNIST dataset, we train Lenet-5 network by using SGD with $L_{1}-$norm
and $L_{2}-$norm on parameters. In contrast, we also employ our $S^{2}$-LBI
to train the Lenet-5 structure. We visualize the conv.c5 filters of
LeNet-5 in Fig. \ref{path_visualization}. We visualize the solution
path in the left column, by comparing the changing of the magnitude
of each filter with the training epochs. We use red color to represent
the filters that have learned more obvious visual patterns. Particularly,
in our $S^{2}$-LBI algorithm, the solution path of filters selected
in $\widetilde{W}$ are drawn by red color. Thus, it is quite clear
that both $S^{2}$-LBI and SGD with $L_{1}-$norm could produce a
sparse DNN structure, and the magnitudes of most of filters are very
small.

\section{Conclusion}

This paper proposes a $S^{2}$-LBI algorithm to train and boost the
network, important filters at the same time. Particularly, the proposed
algorithm, for the first time, can automatically grow or simplify a network.
The experiments show
the efficacy of proposed framework.

\bibliography{iclr2019_conference}

\begin{thebibliography}{47}
\providecommand{\natexlab}[1]{#1}
\providecommand{\url}[1]{\texttt{#1}}
\expandafter\ifx\csname urlstyle\endcsname\relax
  \providecommand{\doi}[1]{doi: #1}\else
  \providecommand{\doi}{doi: \begingroup \urlstyle{rm}\Url}\fi

\bibitem[Abbasi-Asl and Yu(2017)]{greedy_filter_pruning}
Reze Abbasi-Asl and Bin Yu.
\newblock Structural compression of convolutional neural networks based on
  greedy filter pruning.
\newblock In \emph{arxiv}, 2017.

\bibitem[Alvarez and Salzmann(2016)]{group_spars_network}
Jose~M Alvarez and Mathieu Salzmann.
\newblock Learning the number of neurons in deep networks.
\newblock In \emph{NIPS}, 2016.

\bibitem[Bottou(2010)]{bottou-2010}
L\'{e}on Bottou.
\newblock Large-scale machine learning with stochastic gradient descent.
\newblock In \emph{COMPSTAT}, 2010.

\bibitem[B\"uhlmann and Yu(2003)]{B2003Boosting}
Peter B\"uhlmann and Bin Yu.
\newblock Boosting with the l2 loss: Regression and classification.
\newblock \emph{Journal of the American Statistical Association}, 98\penalty0
  (462):\penalty0 324--339, 2003.

\bibitem[Collins and Kohli(2014)]{l1_memory}
Maxwell Collins and Pushmeet Kohli.
\newblock Memory bounded deep convolutional networks.
\newblock In \emph{arXiv preprint arXiv:1412.1442, 2014}, 2014.

\bibitem[Elsken et~al.(2018)Elsken, Metzen, and Hutter]{nas_survey_2018}
Thomas Elsken, Jan~Hendrik Metzen, and Frank Hutter.
\newblock Neural architecture search: A survey.
\newblock In \emph{arxiv:1808.05377}, 2018.

\bibitem[Erhan et~al.(2009)Erhan, Bengio, Courville, and
  Vincent]{Erhan2009Visualizing}
Dumitru Erhan, Yoshua Bengio, Aaron Courville, and Pascal Vincent.
\newblock Visualizing higher-layer features of a deep network.
\newblock \emph{University of Montreal}, 1341\penalty0 (3):\penalty0 1, 2009.

\bibitem[Han et~al.(2015{\natexlab{a}})Han, Mao, and Dally]{han2015deep}
Song Han, Huizi Mao, and William~J Dally.
\newblock Deep compression: Compressing deep neural networks with pruning,
  trained quantization and huffman coding.
\newblock In \emph{ICLR}, 2015{\natexlab{a}}.

\bibitem[Han et~al.(2015{\natexlab{b}})Han, Pool, Tran, and
  Dally]{han2015learning}
Song Han, Jeff Pool, John Tran, and William Dally.
\newblock Learning both weights and connections for efficient neural network.
\newblock In \emph{NIPS}, 2015{\natexlab{b}}.

\bibitem[He et~al.(2015)He, Zhang, Ren, and Sun]{he2015delving}
Kaiming He, Xiangyu Zhang, Shaoqing Ren, and Jian Sun.
\newblock Delving deep into rectifiers: Surpassing human-level performance on
  imagenet classification.
\newblock In \emph{ICCV}, 2015.

\bibitem[He et~al.(2016)He, Zhang, Ren, and Sun]{he2016deep}
Kaiming He, Xiangyu Zhang, Shaoqing Ren, and Jian Sun.
\newblock Deep residual learning for image recognition.
\newblock In \emph{Proceedings of the IEEE conference on computer vision and
  pattern recognition}, pages 770--778, 2016.

\bibitem[Hinton et~al.(2014)Hinton, Vinyals, and Dean]{hinton_distill}
Geoffrey Hinton, Oriol Vinyals, and Jeff Dean.
\newblock Distilling the knowledge in a neural network.
\newblock In \emph{NIPS 2014 Deep Learning Workshop}, 2014.

\bibitem[Howard et~al.(2017)Howard, Zhu, Chen, Kalenichenko, Wang, Weyand,
  Andreetto, and Adam]{howard2017mobilenets}
Andrew~G. Howard, Menglong Zhu, Bo~Chen, Dmitry Kalenichenko, Weijun Wang,
  Tobias Weyand, Marco Andreetto, and Hartwig Adam.
\newblock Mobilenets: Efficient convolutional neural networks for mobile vision
  applications.
\newblock In \emph{arxiv}, 2017.

\bibitem[Huang et~al.(2016{\natexlab{a}})Huang, Sun, Xiong, and Yao]{Splitlbi}
Chendi Huang, Xinwei Sun, Jiechao Xiong, and Yuan Yao.
\newblock Split lbi: An iterative regularization path with structural sparsity.
  advances in neural information processing systems.
\newblock \emph{Advances In Neural Information Processing Systems}, pages
  3369--3377, 2016{\natexlab{a}}.

\bibitem[Huang et~al.(2016{\natexlab{b}})Huang, Sun, Xiong, and
  Yao]{huang2016split}
Chendi Huang, Xinwei Sun, Jiechao Xiong, and Yuan Yao.
\newblock Split lbi: An iterative regularization path with structural sparsity.
\newblock In \emph{NIPS}, pages 3369--3377, 2016{\natexlab{b}}.

\bibitem[Huang et~al.(2018)Huang, Sun, Xiong, and Yuan]{Huang2017Boosting}
Chendi Huang, Xinwei Sun, Jiechao Xiong, and Yao Yuan.
\newblock Boosting with structural sparsity: A differential inclusion approach.
\newblock \emph{Applied and Computational Harmonic Analysis}, 2018.

\bibitem[Iandola et~al.(2017)Iandola, Han, Moskewicz, Ashraf, Dally, and
  Keutzer]{squeezenet}
Forrest~N. Iandola, Song Han, Matthew~W. Moskewicz, Khalid Ashraf, William~J.
  Dally, and Kurt Keutzer.
\newblock Squeezenet: Alexnet-level accuracy with 50x fewer parameters and
  <0.5mb model size.
\newblock In \emph{ICLR}, 2017.

\bibitem[Jaderberg et~al.(2014)Jaderberg, Vedaldi, and
  Zisserman]{jaderberg2014speeding}
Max Jaderberg, Andrea Vedaldi, and Andrew Zisserman.
\newblock Speeding up convolutional neural networks with low rank expansions.
\newblock In \emph{BMVC}, 2014.

\bibitem[Kingma and Ba(2015)]{kingma2014adam}
Diederik Kingma and Jimmy Ba.
\newblock Adam: A method for stochastic optimization.
\newblock In \emph{ICLR}, 2015.

\bibitem[Krichene et~al.(2015)Krichene, Bayen, and
  Bartlett]{krichene2015accelerated}
Walid Krichene, Alexandre Bayen, and Peter~L Bartlett.
\newblock Accelerated mirror descent in continuous and discrete time.
\newblock In \emph{Advances in neural information processing systems}, pages
  2845--2853, 2015.

\bibitem[LeCun et~al.(1998)LeCun, Bottou, Bengio, and Haffner]{LeCunCNN}
Yann LeCun, Leon Bottou, Yoshua Bengio, and Patrick Haffner.
\newblock Gradient-based learning applied to document recognition.
\newblock 86\penalty0 (11):\penalty0 2278--2324, 1998.

\bibitem[Li et~al.(2017)Li, Kadav, Durdanovic, Samet, and Graf]{li2016pruning}
Hao Li, Asim Kadav, Igor Durdanovic, Hanan Samet, and Hans~Peter Graf.
\newblock Pruning filters for efficient convnets.
\newblock In \emph{ICLR}, 2017.

\bibitem[Li and Hoiem(2016)]{zhizhong2016eccv}
Zhizhong Li and Derek Hoiem.
\newblock Learning without forgetting.
\newblock In \emph{ECCV}, 2016.

\bibitem[Liu et~al.(2017)Liu, Li, Shen, Huang, Yan, and Zhang]{Liu_2017_ICCV}
Zhuang Liu, Jianguo Li, Zhiqiang Shen, Gao Huang, Shoumeng Yan, and Changshui
  Zhang.
\newblock Learning efficient convolutional networks through network slimming.
\newblock In \emph{ICCV}, 2017.

\bibitem[Ma et~al.(2018)Ma, Zhang, Zheng, and Sun]{shufflenetv2}
Ningning Ma, Xiangyu Zhang, Hai-Tao Zheng, and Jian Sun.
\newblock Shufflenet v2: Practical guidelines for efficient cnn architecture
  design.
\newblock In \emph{arXiv:1807.11164v1}, 2018.

\bibitem[Mason et~al.(1999)Mason, Baxter, Bartlett, and
  Frean]{Mason1999boosting}
Llew Mason, Jonathan Baxter, Peter Bartlett, and Marcus Frean.
\newblock Boosting algorithms as gradient descent.
\newblock In \emph{International Conference on Neural Information Processing
  Systems}, 1999.

\bibitem[Molchanov et~al.(2017)Molchanov, Tyree, Karras, Aila, and
  Kautz]{molchanov2016pruning}
Pavlo Molchanov, Stephen Tyree, Tero Karras, Timo Aila, and Jan Kautz.
\newblock Pruning convolutional neural networks for resource efficient transfer
  learning.
\newblock In \emph{ICLR}, 2017.

\bibitem[Osher et~al.(2005)Osher, Burger, Goldfarb, Xu, and
  Yin]{osher2005iterative}
Stanley Osher, Martin Burger, Donald Goldfarb, Jinjun Xu, and Wotao Yin.
\newblock An iterative regularization method for total variation-based image
  restoration.
\newblock \emph{Multiscale Modeling \& Simulation}, 4\penalty0 (2):\penalty0
  460--489, 2005.

\bibitem[Osher et~al.(2016)Osher, Ruan, Xiong, Yao, and Yin]{bregman}
Stanley Osher, Feng Ruan, Jiechao Xiong, Yuan Yao, and Wotao Yin.
\newblock Sparse recovery via differential inclusions.
\newblock \emph{Applied and Computational Harmonic Analysis}, 2016.

\bibitem[Pentina and Lampert(2015)]{lifelong_iid}
Anastasia Pentina and Christoph~H. Lampert.
\newblock Lifelong learning with non-i.i.d. tasks.
\newblock In \emph{NIPS}. 2015.

\bibitem[Ranzato and Hinton(2010)]{Hinton2010cvpr}
M.~Ranzato and G.~E. Hinton.
\newblock Modeling pixel means and covariances using factorized third-order
  boltzmann machines.
\newblock In \emph{CVPR}, 2010.

\bibitem[Ranzato et~al.(2010)Ranzato, Krizhevsky, and Hinton]{Ranzato2010}
M.~Ranzato, A.~Krizhevsky, and G.~E. Hinton.
\newblock Factored 3-way restricted boltzmann machines for modeling natural
  images.
\newblock In \emph{AISTATS}, 2010.

\bibitem[Srivastava et~al.(2014)Srivastava, Hinton, Krizhevsky, Sutskever, and
  Salakhutdinov]{srivastava2014dropout}
Nitish Srivastava, Geoffrey Hinton, Alex Krizhevsky, Ilya Sutskever, and Ruslan
  Salakhutdinov.
\newblock Dropout: a simple way to prevent neural networks from overfitting.
\newblock \emph{The Journal of Machine Learning Research}, 15\penalty0
  (1):\penalty0 1929--1958, 2014.

\bibitem[Sun et~al.(2017)Sun, Hu, Yao, and Wang]{sun2017gsplit}
Xinwei Sun, Lingjing Hu, Yuan Yao, and Yizhou Wang.
\newblock Gsplit lbi: Taming the procedural bias in neuroimaging for disease
  prediction.
\newblock In \emph{International Conference on Medical Image Computing and
  Computer-Assisted Intervention}, pages 107--115. Springer, 2017.

\bibitem[Thrun and Mitchell(1995)]{Tom1995lifelong}
Sebastian Thrun and Tom~M. Mitchell.
\newblock Lifelong robot learning.
\newblock \emph{Robotics and Autonomous Systems}, 1995.

\bibitem[Wang et~al.(2017)Wang, Ramanan, and Hebert]{Wang2017Growing}
Yuxiong Wang, Deva Ramanan, and Martial Hebert.
\newblock Growing a brain: Fine-tuning by increasing model capacity.
\newblock In \emph{CVPR}, 2017.

\bibitem[Wen et~al.(2016{\natexlab{a}})Wen, Wu, Wang, Chen, and Li]{l12_norm}
Wei Wen, Chunpeng Wu, Yandan Wang, Yiran Chen, and Hai Li.
\newblock Learning the number of neurons in deep networks.
\newblock In \emph{NIPS}, 2016{\natexlab{a}}.

\bibitem[Wen et~al.(2016{\natexlab{b}})Wen, Wu, Wang, Chen, and
  Li]{wen2016learning}
Wei Wen, Chunpeng Wu, Yandan Wang, Yiran Chen, and Hai Li.
\newblock Learning structured sparsity in deep neural networks.
\newblock In \emph{NIPS}, 2016{\natexlab{b}}.

\bibitem[Yang et~al.(2018)Yang, Kang, Dong, Fu, and Yang]{soft_filtering}
He~Yang, Guoliang Kang, Xuanyi Dong, Yanwei Fu, and Y.~Yang.
\newblock Soft filter pruning for accelerating deep convolutional neural
  networks.
\newblock In \emph{IJCAI 2018}, 2018.

\bibitem[Yoon and Hwang(2017)]{yoon2017combined}
Jaehong Yoon and Sung~Ju Hwang.
\newblock Combined group and exclusive sparsity for deep neural networks.
\newblock In \emph{ICML}, 2017.

\bibitem[Yuan and Lin(2006)]{yuan2006model}
Ming Yuan and Yi~Lin.
\newblock Model selection and estimation in regression with grouped variables.
\newblock \emph{Journal of the Royal Statistical Society: Series B (Statistical
  Methodology)}, 68\penalty0 (1):\penalty0 49--67, 2006.

\bibitem[Yuan et~al.(2007)Yuan, Rosasco, and Caponnetto]{Yuan2007On}
Yao Yuan, Lorenzo Rosasco, and Andrea Caponnetto.
\newblock On early stopping in gradient descent learning.
\newblock \emph{Constructive Approximation}, 26\penalty0 (2):\penalty0
  289--315, 2007.

\bibitem[Zhang et~al.(2016)Zhang, Zou, He, and Sun]{zhang2016accelerating}
Xiangyu Zhang, Jianhua Zou, Kaiming He, and Jian Sun.
\newblock Accelerating very deep convolutional networks for classification and
  detection.
\newblock \emph{IEEE Transactions on Pattern Analysis and Machine
  Intelligence}, 38\penalty0 (10):\penalty0 1943--1955, 2016.

\bibitem[Zhao et~al.(2018)Zhao, Sun, Fu, Yao, and Wang]{zhao2018msplit}
Bo~Zhao, Xinwei Sun, Yanwei Fu, Yuan Yao, and Yizhou Wang.
\newblock Msplit lbi: Realizing feature selection and dense estimation
  simultaneously in few-shot and zero-shot learning.
\newblock \emph{arXiv preprint arXiv:1806.04360}, 2018.

\bibitem[Zhou et~al.(2017)Zhou, Yao, Guo, Xu, and Chen]{zhou2017incremental}
Aojun Zhou, Anbang Yao, Yiwen Guo, Lin Xu, and Yurong Chen.
\newblock Incremental network quantization: Towards lossless cnns with
  low-precision weights.
\newblock \emph{ICLR}, 2017.

\bibitem[Zhu et~al.(2017)Zhu, Han, Mao, and Dally]{zhu2016trained}
Chenzhuo Zhu, Song Han, Huizi Mao, and William~J Dally.
\newblock Trained ternary quantization.
\newblock \emph{ICLR}, 2017.

\bibitem[Zoph and Le(2016)]{zoph2016neural}
Barret Zoph and Quoc~V Le.
\newblock Neural architecture search with reinforcement learning.
\newblock \emph{arXiv preprint arXiv:1611.01578}, 2016.

\end{thebibliography}

\end{document}